\documentclass[journal]{IEEEtran}
\usepackage{amsmath,amsfonts,amssymb}
\usepackage{array}
\usepackage[caption=false,font=normalsize,labelfont=sf,textfont=sf]{subfig}
\usepackage{textcomp}
\usepackage{stfloats}
\usepackage{url}
\usepackage{verbatim}
\usepackage{graphicx}
\usepackage{cite}
\usepackage{setspace}
\usepackage{cite}
\usepackage{hyperref}
\usepackage{tabularx}
\usepackage[utf8]{inputenc}
\usepackage{times}  
\usepackage{helvet}  
\usepackage{courier}  
\usepackage[ruled,lined,linesnumbered]{algorithm2e}
\DontPrintSemicolon
\SetKwComment{tcc}{$\triangleright$~}{}
\SetCommentSty{normalfont}
\SetKwInput{KwRequire}{Require}
\SetNlSty{}{}{:}
\usepackage{paralist}
\usepackage[most]{tcolorbox}
\usepackage{todonotes}
\usepackage[capitalise]{cleveref}
\usepackage[legacycolonsymbols]{mathtools}
\newtheorem{definition}{Definition}

\usepackage{booktabs}
\usepackage{ifthen}
\usepackage{xr}
\graphicspath{ {./fig/} }

\begin{document}

\title{Variable Time Step Reinforcement Learning for Robotic Applications}

\author{Dong Wang,~\IEEEmembership{Member,~IEEE}, Giovanni
  Beltrame,~\IEEEmembership{Senior Member,~IEEE}%
  \thanks{Manuscript received ..., 2024; accepted ,,,. This paper was recommended for
    publication by Editor XXX upon evaluation of the
    reviewers’ comments. This work was partly presented at the Finding the Frame
    Workshop in August 2024.} \thanks{D. Wang and G. Beltrame are with the Department of
    Computer and Software Engineering, Polytechnique Montr\'eal, Montr\'eal, QC,
    H3T1J4 Canada e-mail: \{dong-1.wang,giovanni.beltrame\}@polymtl.ca
  }
}

\markboth{TRANSACTIONS ON ROBOTICS,~Vol.~xx, No.~x, date}{Dong
  \MakeLowercase{\textit{et al.}}}%
\maketitle

%

\IEEEpeerreviewmaketitle

\begin{abstract}
  Traditional reinforcement learning (RL) generates discrete control policies,
  assigning one action per cycle. These policies are usually implemented as in a
  fixed-frequency control loop. This rigidity presents challenges as optimal
  control frequency is task-dependent; suboptimal frequencies increase
  computational demands and reduce exploration efficiency. Variable Time Step
  Reinforcement Learning (VTS-RL) addresses these issues with adaptive control
  frequencies, executing actions only when necessary, thus reducing
  computational load and extending the action space to include action durations.
  In this paper we introduce the Multi-Objective Soft Elastic Actor-Critic
  (MOSEAC) method to perform VTS-RL, validating it through theoretical analysis
  and experimentation in simulation and on real robots. Results show faster
  convergence, better training results, and reduced energy consumption with
  respect to other variable- or fixed-frequency approaches.
\end{abstract}

\begin{IEEEkeywords}
Variable Time Step Reinforcement Learning (VTS-RL), Deep Learning in Robotics 
and Automation, Learning and Adaptive Systems, Optimization and Optimal Control
\end{IEEEkeywords}

\section{Introduction}
\IEEEPARstart{D}{eep} reinforcement learning (DRL) algorithms have achieved
significant success in gaming \cite{liu2021deep, ibarz2021train} and robotic
control \cite{akalin2021reinforcement, singh2022reinforcement}. Traditional DRL
employs a fixed control loop at set intervals (e.g., every 0.1 seconds). This
fixed-rate control can cause stability issues and high computational demands,
particularly in dynamic environments where the optimal action frequency varies.

Variable Time Step Reinforcement Learning (VTS-RL) has been recently proposed to
address these issues. Based on reactive programming principles
\cite{majumdar2021paracosm, bregu2016reactive}, VTS-RL executes control actions
only when necessary, reducing computational load and expanding the action space
to include variable action durations. For example, in robotic manipulation,
VTS-RL allows a robot arm to dynamically adjust its control frequency, using
lower frequencies during simple, repetitive tasks and higher frequencies during
complex maneuvers or when handling delicate objects \cite{karimi2023dynamic}.

Two notable VTS-RL algorithms are Soft Elastic Actor-Critic (SEAC)
\cite{wang2024deployable} and Continuous-Time Continuous-Options (CTCO)
\cite{karimi2023dynamic}. CTCO employs continuous-time decision-making with
flexible option durations. CTCO requires tuning several hyperparameters, such as
its radial basis functions (RBFs) and the time-related hyperparameters $\tau$
for its adaptive discount factor $\gamma$, complicating tuning in some
environments.

SEAC incorporates reward terms for task energy (number of actions) and task
time, making it effective in time-restricted environments like racing video
games \cite{wang2024reinforcement}. However, it also requires careful
hyperparameter tuning (balancing task, energy, and time costs) to maintain
performance. The sensitivity of SEAC and CTCO to hyperparameter settings
challenges users to fully leverage their potential.

We introduce the Multi-Objective Soft Elastic Actor-Critic (MOSEAC)
algorithm \cite{wang2024moseac}. MOSEAC integrates action durations into the
action space and adjusts hyperparameters based on observed trends in task
rewards during training, reducing the need to set multiple hyperparameters.
Additionally, our hyperparameter setting approach can be broadly applied to any
continuous action reinforcement learning algorithm.
This adaptability facilitates the transition from fixed-time step to
variable-time step reinforcement learning, significantly expanding its practical
application.


In this paper, we provide an in-depth analysis of MOSEAC's theoretical
performance, covering its framework, implementation, performance guarantees,
convergence and complexity analysis. We also deploy the MOSEAC model as the
navigation policy in simulation and on a physical AgileX Limo
\cite{agilex_limo}. Our evaluation includes statistical performance analysis,
trajectory similarity analysis between simulated and real environments, and a
comparison of average computational resource consumption on both CPU and GPU.

MOSEAC allows the optimization of the control frequency for RL policies, as well
as a reduction in computational load on the robots' onboard computer. The
resources saved can be redirected to other critical tasks such as environmental
sensing \cite{lajoie2020door} and communication \cite{wan2020cognitive}.
In addition, the MOSEAC principles can be applied to a variety of RL algorithms,
making it a useful tool for the deployment of RL policies on physical robotics
systems, reducing the real-to-sim gap.


The paper is organized as follows. Section~\ref{sec:related} reviews the 
current research status of VTS-RL. Section~\ref{sec:algorithm} 
details the MOSEAC framework with its pseudocode. Section~\ref{sec:theoreties} 
presents the theoretical analysis of MOSEAC. Section~\ref{sec:experiment} 
describes the implementation of our validation environment in the real world 
and the method to build the simulation environment. Section~\ref{sec:result} 
presents the experiment results on both the simulation and the real Agilex 
Limo. Finally, Section~\ref{sec:conclusions} concludes the paper.

\section{Related Work}
\label{sec:related}

The importance of action duration in reinforcement learning (RL) has been
significantly underestimated, yet it is crucial for applying RL algorithms in
real-world scenarios, impacting an agent's exploration capabilities. For
instance, high frequencies may reduce exploration efficiency but are essential
in delayed environments \cite{bouteiller2021reinforcement}. Recent studies by
Amin et al. \cite{amin2020locally}, and Park et al. \cite{park2021time} have
highlighted this issue, showing that variable action durations can significantly
affect learning performance. Building on these insights, Wang \&
Beltrame~\cite{wang2024reinforcement} and Karimi et al.~\cite{karimi2023dynamic}
further explored how different frequencies impact learning, noting that
excessively high frequencies can impede convergence. Their findings suggest that
dynamic control frequencies, which adjust in real-time based on reactive
principles, could enhance performance and adaptability.

Expanding on the idea of adaptive control, Sharma et
al.~\cite{sharma2017learning} introduced a related concept of repetitive action
reinforcement learning, where an agent performs identical actions over
successive states, combining them into a larger action. This concept was further
explored by Metelli et al.~\cite{metelli2020control} and Lee et
al.~\cite{lee2020reinforcement} in gaming contexts. However, despite its
potential, this method does not fully address physical properties or
computational demands, limiting its practical application.

Additionally, Chen et al.~\cite{chen2021varlenmarl} proposed modifying the
traditional control rate by integrating actions such as "sleep" to reduce
activity periods. However, this approach still required fixed-frequency system
checks, ultimately failing to reduce computational load as intended.

In the domain of traffic signal control, research focused on hybrid action
reinforcement learning. The study highlighted the importance of synchronously
optimizing stage specifications and green interval durations, underscoring the
potential of variable action durations to improve decision-making processes in
complex, real-time environments~\cite{even2003action}.

Furthermore, research on variable damping control for wearable robotic limbs has
showcased the application of reinforcement learning to adjust control parameters
adaptively in response to changing states. This study illustrated the
effectiveness of RL in maintaining system stability and performance under
varying conditions, reinforcing the value of dynamic control frequencies in
practical applications~\cite{zhao2023reinforcement}.

These studies collectively highlight the ongoing efforts to integrate variable
action durations and adaptive control mechanisms into RL. They also underscore
the nascent stage of effectively integrating variable control frequencies and
repetitive behaviors into practical applications, emphasizing their critical
importance for enhancing algorithm performance and applicability in real-world
scenarios.

\section{Algorithm Framework}
\label{sec:algorithm}

Soft Elastic Actor-Critic (SEAC) is an extension of the Soft Actor-Critic (SAC)
algorithm that addresses the limitations of fixed control rates in reinforcement
learning (RL) \cite{wang2024deployable}. Traditional Markov Decision Processes
(MDPs) in RL do not account for the duration of actions, assuming that all
actions are executed over uniform time steps. This can lead to inefficiencies,
as the time between two actions can vary widely, requiring a fixed-frequency
control rate in practical deployments. SEAC breaks this assumption by
dynamically adjusting the duration of each action based on the state and
environmental conditions, following the principles of reactive programming. By
incorporating action duration $D$ into the action set, SEAC can decide on both
the action and its duration to optimize energy consumption and computational
efficiency.

MOSEAC extends SEAC~\cite{wang2024deployable} and combines its hyperparameters
for balancing task, energy, and time rewards, and provides a method for
automatically adjust its other hyperparamenters during training.
Similar to SEAC, MOSEAC reward includes components for:
\begin{compactitem}
	\item Quality of task execution (the standard RL reward),
	\item Time required to complete a task (important for varying action durations), and
	\item Energy (the number of time steps, a.k.a. the number of actions taken, which we aim to minimize).
\end{compactitem}

\begin{definition}\label{def:reward_policy}
  The reward associated with the state space is:
	\begin{equation}
	R = \alpha_{m} R_t R_{\tau} - \alpha_{\varepsilon}
	\end{equation}
	where $R_{t}$ is the task reward and $R_{\tau}$ is a time-dependent term.
	
	$\alpha_{m}$ is a weighting factor used to modulate the magnitude of the
  reward: its primary function is to prevent the reward from being too small,
  which could lead to task failure, or too large, which could cause reward
  explosion \cite{Gottipati2020MaxReward}, ensuring stable learning.
	
	$\alpha_{\varepsilon}$ is a penalty parameter applied at each time step to
  impose a cost on the agent’s actions. This parameter gives a fixed cost to the
  execution of an action, thereby discouraging unnecessary ones. In practice,
  $\alpha_{\varepsilon}$ promotes the completion of a task using fewer time
  steps (remember that time steps have variable duration), i.e., it reduces the
  energy used by the control loop of the agent.
	
	We determine the optimal policy $\pi^*$, which maximizes the reward $R$.
\end{definition}

The reward is designed to minimize both energy cost (number of steps) and the
total time to complete the task through $R_{\tau}$. By scaling the task-specific
reward based on action duration with $\alpha_m$, agents are motivated to
complete tasks using fewer actions:

\begin{definition}\label{def:time_reward}
	The remap relationship between action duration and reward is:
	\begin{equation}
	R_{\tau} = D_{min} / D, \quad R_{\tau} \in [D_{min}/D_{max}, 1]
	\end{equation}
	where $t$ is the duration of the current action, $D_{min}$ is the minimum
  duration of an action (strictly greater than 0), and $D_{max}$ is the maximum
  duration of an action.
\end{definition}

We automatically set $\alpha_{m}$ and $\alpha_{\varepsilon}$ during training.
Based on previous results~\cite{wang2024reinforcement}, we bind the increase of
$\alpha_{m}$ to a decrease $\alpha_{\varepsilon}$ using a sigmoid function to
mitigate convergence issues, specifically the problem of sparse rewards caused a
large $\alpha_{\varepsilon}$ and a small $\alpha_{m}$. This adjustment ensures
that rewards are appropriately balanced, facilitating learning and convergence.

\begin{definition}\label{def:r_e}
	The relationship between $\alpha_{\varepsilon}$ and $\alpha_{m}$ is:
	\begin{equation}
	\alpha_{\varepsilon} = 0.2 \cdot \left(1 - \frac{1}{1 + e^{-\alpha_m + 1}}\right)
	\end{equation}
	Based on \cite{wang2024deployable}'s experience, we establish a mapping
  relationship between the two parameters: when the initial value of
  $\alpha_{m}$ is 1.0, the initial value of $\alpha_{\varepsilon}$ is 0.1. As
  $\alpha_{m}$ increases, $\alpha_{\varepsilon}$ decreases, but never falls
  below 0.
\end{definition}

Overall, we update $\alpha_m$ based on the current trend in average reward
during training. To guarantee stability, we ensure the change is monotonic,
forcing a uniform sweep of the parameter space, and a maximum value of
$\alpha_m$, namely $\alpha_{max}$. To determine the trend in the
average reward, we perform a linear regression across the current training
episode and compute the slope of the resulting line: if it is negative, the
reward is declining.

\begin{definition}\label{def:linregress}
	The slope of the average reward ($k_R$) is:
	\begin{equation}
    k_R(R_a) = \frac{n \sum_{i=1}^{n} (i \cdot {R_a}_i) - \left( \sum_{i=1}^{n} 
        i \right) \left( \sum_{i=1}^{n} {R_a}_i \right)}{n \sum_{i=1}^{n} i^2 - 
      \left( \sum_{i=1}^{n} i \right)^2}
	\end{equation}
	where $n$ is the total number of data points collected across the update
  interval ($k_{update}$ in Algorithm \autoref{algo:SEAC}). The update interval
  is a hyperparameter that determines the frequency of updates for these neural
  networks used in the actor and critic policies, occurring after every $n$
  episode \cite{sutton2018reinforcement}. $R_a$ represents the list of average
  rewards $({R_a}_1,{R_a}_2,...,{R_a}_n)$ during training. Here, an average
  reward ${R_{a}}_i$ is calculated across one episode.
\end{definition}

\begin{definition}\label{def:monotonic}
	We adaptively adjust the reward every $k_{update}$ episodes when $k_R < 0$:
	\begin{equation}
	\left\{
	\begin{array}{ll}
	\alpha_{m} = \alpha_{m} + \psi & \text{if } \alpha_{m} < \alpha_{max} \\
	\alpha_{m} = \alpha_{max} & \text{otherwise}
	\end{array}
	\right.
	\end{equation}
	where $\psi$ serves as the sole additional MOSEAC hyperparameter necessary for
  adjusting the reward equation during training. The parameter $\alpha_{max}$
  represents the upper limit of $\alpha_{m}$, ensuring algorithmic convergence
  and preventing reward explosion. Furthermore, $\alpha_{\varepsilon}$ is
  adjusted in accordance with Definition \autoref{def:r_e}.
\end{definition}


Algorithm \autoref{algo:SEAC} shows the pseudocode of MOSEAC. In short, MOSEAC
extends the SAC algorithm by incorporating action duration $D$ into the action
policy set. This expansion allows the algorithm to predict the action and its
duration simultaneously. The reward is calculated using
\autoref{def:reward_policy}, and its changes are continuously monitored. If the
reward trend declines, $\alpha_m$ increases linearly at a rate of $\psi$,
without exceeding $alpha_{max}$. The action and critic networks are periodically
updated like the SAC algorithm based on these preprocessed rewards.

\begin{algorithm}[htbp]\label{algo:SEAC}
	\SetAlgoLined
	\KwRequire{a policy $\pi$ with a set of parameters $\theta$, $\theta^{'}$, 
		critic parameters $\phi$, $\phi^{'}$, variable time step environment 
		model $\Omega$, learning-rate $\lambda_p$, $\lambda_q$, reward buffer 
		$\beta_r$, replay buffer $\beta$.}
	
	Initialization $i = 0$, $t_i = 0$, $\beta_r = 0$, observe $S_0$\\
	\While{$t_i \leq t_{max}$}{
		\For{$i \leq k_{length} \vee Not \, Done$}{
			$A_i, D_i = \pi_{\theta}(S_{i})$ \\ \hfill {$\rightarrow$ simple 
				action and its duration}\\
			$S_{i+1}, R_i = \Omega(A_i, D_i)$ \\ \hfill {$\rightarrow$ compute 
				reward with Definition \ref{def:reward_policy} and 
				\ref{def:time_reward}}\\
			
			$i \leftarrow i+1$
		}
		$\beta_{r} \leftarrow 1/i \times \sum_{0}^{i} R_i$ \\ \hfill 
		{$\rightarrow$ collect the average reward for one episode}\\
		$\beta \leftarrow S_{0 \sim i}, \, A_{0 \sim i}, \, D_{0 \sim i}, \, 
		R_{0 \sim i}, \, S_{1 \sim i+1}$ \\
		$i = 0$\\
		$t_{i} \leftarrow t_{i} + 1$ \\ 
		\If{$t_{i} \geq k_{init} \quad \& \quad t_{i} \mid k_{update}$}{
			$Sample \, S, \, A, \, D, \, R, \, S^{'} from (\beta)$ \\
			$\phi \leftarrow \phi - \lambda_q\nabla_{\delta}\mathcal{L}_{Q}
			(\phi, \, S, \, A, \, D, \, R, \, S^{'})$ \\ \hfill {$\rightarrow$ 
				critic update}\\
			$\theta \leftarrow \theta - \lambda_p\nabla_{\theta}\mathcal{L}_{
				\pi}(\theta, \, S, \, A, \, D, \, \phi)$ \\ \hfill 
			{$\rightarrow$ actor update}\\
			\If{$k_R(\beta_r)$}{
				
				$\alpha_{m} = \alpha_{m} + \psi$  \hfill $\text{if} \alpha_{m} 
				< \alpha_{max}$ \\
				$\text{Or}$ $\alpha_{m} = \alpha_{max}$  \hfill 
				$\text{otherwise}$ \\
				 \hfill {$\rightarrow$ see Definition \autoref{def:linregress} 
				 	for $k_R$}\\
				$\alpha_{\varepsilon} \leftarrow F_{update}(\alpha_{m})$ \\ 
				\hfill {$\rightarrow$ update $\alpha_{m}, \alpha_{\varepsilon}$ 
					followed Definition \ref{def:r_e}}\\}
			$\beta_{r} = 0$ \\ \hfill {$\rightarrow$ Re-record average reward 
				values under new hyperparameters}\\
		}
		Perform soft-update of $\phi^{'}$ and $\theta^{'}$
	}
	\caption{Multi-Objective Soft Elastic Actor and Critic (MOSEAC)}
\end{algorithm}

Here, $t_{max}$ represents the maximum number of training steps
\cite{sutton2018reinforcement}; $k_{length}$ is the maximum number of
exploration steps per episode \cite{sutton2018reinforcement}; $k_{init}$ is the
number of steps in the initial random exploration phase
\cite{sutton2018reinforcement}. The reward $R(S_i, A_i, D_i)$ depends on state
($S_i$), action ($A_i$) and duration $D_i \in [D_{min}, D_{max}]$.

Overall, our reward scalarizes a multi-objective optimization problem including
rewards for task, time, and energy. Unlike Hierarchical Reinforcement Learning
(HRL) \cite{dietterich2000hierarchical, li2019hierarchical}, which seeks Pareto
optimality \cite{kacem2002pareto, monfared2021pareto} with layered reward
policies, our method simplifies the approach, making our strategy easily
adaptable to various algorithms. While our goal is to avoid the complexity of
Pareto optimization, dynamically adjusting the reward structure inevitably
places us on the Pareto optimization curve. For a theoretical analysis of the
Pareto curve and how $\alpha_{max}$ ensures algorithmic stability, please refer
to Appendix~\ref{appendix:adaptive}.

Apart from a series of hyperparameters inherent to RL that need adjustment, such
as learning rate, $t_{max}$, $k_{update}$, etc., $\psi$ is the primary
hyperparameter that requires tuning in MOSEAC.
Determining the appropriate $\psi$ value remains a critical optimization point.
We recommend using the pre-set $\psi$ value provided in our implementation to
reduce the need for additional adjustments. If training performance is poor, a
large $\psi$ may cause the reward signal's gradient to change too rapidly,
leading to instability, suggesting a reduction in $\psi$. Conversely, if
training is slow, a small $\psi$ may result in a weak reward signal, affecting
convergence, suggesting an increase in $\psi$.

\section{Theoretical Analysis}
\label{sec:theoreties}
We have analyzed the theoretical performance of MOSEAC through various aspects,
including performance guarantees, convergence and complexity analysis.
\autoref{table:symbol_notation} provides the notation for the following.

\subsection{Performance Guarantees}
\label{sec:performance}

	
In the standard SAC algorithm \cite{haarnoja2018soft}, the policy
$\pi_\theta(a|s)$ selects action $a$, and updates the policy parameters $\theta$
and value function parameters $\phi$. The objective function is:
\begin{equation}
	J(\pi_\theta) = \mathbb{E}_{(s, a) \sim \pi_\theta} \left[ Q^\pi(s, a) +
    \alpha \mathcal{H}(\pi_\theta(\cdot|s)) \right]
\end{equation}
where $J(\pi_\theta)$ is the objective function, $Q^\pi(s, a)$ is the
state-action value function, $\alpha$ is a temperature parameter controlling the
entropy term, and $\mathcal{H}(\pi_\theta(\cdot|s))$ is the entropy of the
policy.
	
When the action space is extended to include the action duration $D$ and the
reward function is modified to $R = \alpha_{m} R_{t} R_{\tau} -
\alpha_{\varepsilon}$, where $\alpha_{m} \geq 0$, $0 < R_{\tau} \leq 1$, and
$\alpha_{\varepsilon}$ is a small positive constant, the new objective function
becomes:
\begin{equation}
	\begin{aligned}
    J(\pi_\theta) = \mathbb{E}_{(s, a, D) \sim \pi_\theta} \big[ 
    &Q^\pi(s, a, D) + \alpha \mathcal{H}(\pi_\theta(\cdot|s)) \big]
	\end{aligned}
\end{equation}
where $Q^\pi(s, a, D)$ is the extended state-action value function with time
dimension $D$, and $R_{t}$ and $R_{\tau}$ are components of the reward function,
see Definition \autoref{def:reward_policy}.

Replacing $Q^\pi$ the objective function for MOSEAC, incorporating the extended action space 
	and the adaptive reward function, becomes:
	\begin{equation}
    \begin{aligned}
      J(\pi) = \mathbb{E}_{(s_t, a_t, D_t) \sim \rho_{\pi}} \bigg[ &
                                                                     \sum_{t=0}^{\infty} \gamma^t \Big( \alpha_m(t) R_t \frac{D_{\min}}
                                                                     {D_t} \\
                                                                   &- \alpha_\varepsilon(t) + \alpha \mathcal{H}(\pi(\cdot|s_t)) \Big) 
                                                                     \bigg]
    \end{aligned}
	\end{equation}
	where:
	\begin{itemize}
  \item \(\alpha_m(t)\) is the adaptive parameter that increases monotonically
    when the average reward decreases, with an upper limit of \(\alpha_{max}\).
  \item \(\alpha_\varepsilon(t)\) is the adaptive parameter that decreases
    monotonically when the average reward decreases.
  \item \(\mathcal{H}(\pi(\cdot|s_t)) = - \mathbb{E}_{(a_t, D_t) \sim
      \pi(\cdot|s_t)} [\log \pi(a_t, D_t|s_t)]\) is the entropy of the policy.
	\end{itemize}
	
The SAC policy improvement theorem \cite{haarnoja2018soft} ensures that
\begin{equation}
	Q^{\pi_k}(s, a, D) \geq Q^{\pi_{k-1}}(s, a, D)
\end{equation}
for any iteration \( k \geq 0 \).

	
The soft Bellman equation in MOSEAC is:
\begin{equation}
	\begin{aligned}
    Q^{\pi}(s, a, D) &= \alpha_m(t) R_t \frac{D_{\min}}{D} - 
                       \alpha_\varepsilon(t) \\
                     &\quad + \gamma \mathbb{E}_{s' \sim P(\cdot|s, a, D)} \left[ 
                       V^{\pi}(s') \right]
	\end{aligned}
\end{equation}
where the value function \( V^{\pi}(s) \) is defined as:
\begin{equation}
	\begin{aligned}
    V^{\pi}(s) &= \mathbb{E}_{(a, D) \sim \pi(\cdot|s)} \Big[ Q^{\pi}
                 (s, a, D) \\
               &\qquad\quad - \alpha \log \pi(a, D|s) \Big]
	\end{aligned}
\end{equation}	
To prove convergence, we need to show that the soft Bellman operator is a
contraction mapping. Let $\mathcal{T}$ be the soft Bellman operator. For any two
Q-functions $Q_1$ and $Q_2$:
\begin{equation}
  \begin{aligned}
    &\| \mathcal{T}Q_1 - \mathcal{T}Q_2 \|_{\infty} = \sup_{s, a, D} 
			\left| \alpha_m(t) R_t \frac{D_{\min}}{D} - \alpha_\varepsilon(t) 
			\right. \\
    &\quad \left. + \gamma \mathbb{E}_{s' \sim P(\cdot|s, a, D)} \left[ 
			V_1(s') \right] \right. \\
    &\quad \left. - \left( \alpha_m(t) R_t \frac{D_{\min}}{D} - 
			\alpha_\varepsilon(t) + \gamma \mathbb{E}_{s' \sim P(\cdot|s, a, D)}
			\left[ V_2(s') \right] \right) \right| \\
    &\leq \sup_{s, a, D} \gamma \left| \mathbb{E}_{s' \sim 
      P(\cdot|s, a, D)} \left[ V_1(s') - V_2(s') \right] \right| \\
    &\leq \gamma \| V_1 - V_2 \|_{\infty} \\
    &\leq \gamma \| Q_1 - Q_2 \|_{\infty}
  \end{aligned}
\end{equation}
Since $\gamma < 1$, the soft Bellman operator $\mathcal{T}$ is a contraction
mapping. By Banach's fixed-point theorem, there exists a unique fixed point
$Q^*$ such that $Q^* = \mathcal{T}Q^*$~\cite{banach1922operations}.

Let $\pi^*$ be the optimal policy and $\pi_k$ be the policy at iteration $k$.
The error bound between the value functions of the optimal and learned policies
is
\begin{equation}
	\| V^{\pi^*} - V^{\pi_k} \| \leq \frac{2\alpha \gamma \log |\mathcal{A} 
		\times [D_{\min}, D_{\max}]|}{(1 - \gamma)^2}
\end{equation}
With $|\mathcal{A} \times [D_{\min}, D_{\max}]|$ the size of the extended
action space, the contraction property of the soft Bellman operator ensures
that
\begin{equation}
	\| V^{\pi^*} - V^{\pi_k} \|_{\infty} \leq \frac{2\alpha \gamma \log 
		|\mathcal{A} \times [D_{\min}, D_{\max}]|}{(1 - \gamma)^2}
\end{equation}
where $\alpha$ is the coefficient of the entropy term and $\gamma$ is the
discount factor.


\subsection{Convergence Analysis}
\label{sec:convergence_analysis}

With our reward function, the policy gradient is:
\begin{equation}
	\begin{aligned}
		\nabla_\theta J(\pi_\theta) = \mathbb{E}_{\pi_\theta} \Big[ 
		&\nabla_\theta \log \pi_\theta(a, D|s) \cdot \\
		&\big( Q^\pi(s, a, D) \cdot (\alpha_{m} \cdot R_{\tau}) - 
		\alpha_{\varepsilon} \big) \Big]
	\end{aligned}
\end{equation}
where $\nabla_\theta J(\pi_\theta)$ is the gradient of the objective function 
with respect to the policy parameters $\theta$.

The value function update, incorporating the time dimension $D$ and our reward 
function, is:
\begin{equation}
	\begin{aligned}
		L(\phi) = \mathbb{E}_{(s, a, D, r, s')} \Big[ 
		&\big( Q_\phi(s, a, D) - \big( r + \gamma \mathbb{E}_{(a', D') \sim \pi_\theta} \\
		&[V_{\bar{\phi}}(s') - \alpha \log \pi_\theta(a', D'|s')] \big) \big)^2 \Big]
	\end{aligned}
\end{equation}
where $L(\phi)$ is the loss function for the value function update, $r$ is the 
reward, $\gamma$ is the discount factor, and $V_{\bar{\phi}}(s')$ is the target 
value function.

The new policy parameter $\theta$ update rule is:
\begin{equation}
	\begin{aligned}
		\theta_{k+1} = \theta_k + \beta_k \mathbb{E}_{s \sim D, (a, D) \sim \pi_\theta} \Big[ 
		&\nabla_\theta \log \pi_\theta(a, D|s) \cdot \\
		&\big( Q_\phi(s, a, D) \cdot (\alpha_{m} \cdot R_{\tau}) - \alpha_{\varepsilon} \\
		&- V_{\bar{\phi}}(s) + \alpha \log \pi_\theta(a, D|s) \big) \Big]
	\end{aligned}
\end{equation}
where $\beta_k$ is the learning rate at step $k$.

To analyze the impact of dynamically adjusting $\alpha_{m}$ and 
$\alpha_{\varepsilon}$, we assume:
\begin{enumerate}
	\item {\bf{Dynamic Adjustment Rules:}}
		$\alpha_{m}$ increases monotonically by a small increment $\psi$ 
		if the reward trend decreases over consecutive episodes, and its upper 
		limit is $\alpha_{max}$, which guarantees algorithmic convergence and 
		prevents reward explosion. $\alpha_{\varepsilon}$ decreases as defined.
	
	\item {\bf{Learning Rate Conditions:}}
		$\alpha_k$ and $\beta_k$ must satisfy~\cite{konda1999actor}:
		\begin{equation}
			\sum_{k=0}^{\infty} \alpha_k = \infty, \quad \sum_{k=0}^{\infty} \alpha_k^2 < \infty
		\end{equation}
		\begin{equation}
			\sum_{k=0}^{\infty} \beta_k = \infty, \quad \sum_{k=0}^{\infty} \beta_k^2 < \infty
		\end{equation}
\end{enumerate}
Assuming the critic estimates are unbiased:
\begin{equation}
	\mathbb{E}[Q_\phi(s, a, D) \cdot (\alpha_{m} \cdot R_{\tau}) - \alpha_{\varepsilon}] = Q^\pi(s, a, D) \cdot (\alpha_{m} \cdot R_{\tau}) - \alpha_{\varepsilon}
\end{equation}
Since $R_{\tau}$ is a positive number within [0, 1], its effect on 
$Q^\pi(s, a, D)$ is linear and does not affect the consistency of the policy 
gradient.

\begin{enumerate}
\item {\bf{Positive Scaling:}} As $0 \leq R_{\tau} \leq 1$ and $\alpha_{m} \geq
  0$, $\alpha_m$ only scales the reward without altering its sign. This scaling
  does not change the direction of the policy gradient but affects its
  magnitude.
	
\item {\bf{Small Offset:}} $\alpha_{\varepsilon}$ is a small constant used to
  accelerate training. This small offset does not affect the direction of the
  policy gradient but introduces a minor shift in the value function, which does
  not alter the overall policy update direction.
\end{enumerate}
Under these conditions, MOSEAC will converge to a local optimum~\cite{sutton2018reinforcement}:
\begin{equation}
	\lim_{k \to \infty} \nabla_\theta J(\pi_\theta) = 0
\end{equation}

\subsection{Complexity Analysis}
\label{sec:complexity_analysis}

As MOSEAC closely follows SAC~\cite{haarnoja2018soft}, but with an expanded
action set that includes durations, its time complexity is similar to SAC,
replacing the action set $A$ with the expanded one $A \times D$. SAC and MOSEAC
have 3 computational components:
	
	
	
\begin{itemize}
\item{\bf{Policy Evaluation:}} consists of computing the value function
  $V^\pi(s)$ and the Q-value function $Q^\pi(s, a)$ for each state-action pair,
  i.e. $O(|\mathcal{S}| \cdot |\mathcal{A} \times D|)$.
	\item{\bf{Policy Improvement:}} consists updating the policy parameters $\theta$ 
	based on the policy gradient, involving the Q-value function and the 
	entropy term. The time complexity is $O(|\mathcal{S}| \cdot 
	|\mathcal{A} \times D|)$.
\item{\bf{Value Function Update:}} consists of computing the target value using
  the Bellman backup equation, with a time complexity of $O(|\mathcal{S}| \cdot
  |\mathcal{A} \times D|)$.
\end{itemize}
The overall time complexity for one iteration of the SAC algorithm is therefore
\begin{equation}
	O(|\mathcal{S}| \cdot |\mathcal{A} \times D|)	
\end{equation}

The space complexity of MOSEAC is determined by the storage requirements for
the policy, value functions, and other necessary data structures. As for the
time complexity, MOSEAC follows SAC with a different action set:

\begin{itemize}
	\item \textbf{Policy Storage}: Requires $O(|\theta|)$ space for the policy 
	parameters.
	
	\item \textbf{Value Function Storage}: Requires $O(|\mathcal{S}| \cdot 
	|\mathcal{A}| \cdot |D|)$ space for the value functions.
\end{itemize}

The overall space complexity is:
\begin{equation}
O(|\theta| + |\mathcal{S}| \cdot |\mathcal{A}| \cdot |D|)
\end{equation}

The dynamic adjustment of $\alpha_{m}$ and $\alpha_{\varepsilon}$ adds a small 
computational overhead of $O(1)$ per update due to simple arithmetic operations 
and comparisons.

In summary, MOSEAC has higher computational complexity than SAC, but its dynamic
adjustment mechanism and expanded action space provide greater adaptability and
potential for improved performance in multi-objective optimization scenarios.

\begin{table}[h!]
	\centering
	\caption{Complexity Analysis Notation}
	\begin{tabular}{ll}
		\toprule
		Symbol & Description \\
		\midrule
		$\mathcal{S}$ & Set of states \\
		$\mathcal{A}$ & Set of actions \\
		$D$ & Action durations set \\
		$\pi_\theta$ & Policy parameters $\theta$ \\
		$|\theta|$ & Policy parameter count \\
		$V^\pi(s)$ & Value function \\
		$Q^\pi(s, a, D)$ & Q-value function with $D$ \\
		$\alpha$ & Entropy temperature param. \\
		$\alpha_m$ & Reward scaling param. \\
		$\alpha_\epsilon$ & Reward offset param. \\
		$R_t$ & Reward at time \(t\) \\
		$R_{\tau}$ & Time-based reward \\
		$\gamma$ & Discount factor \\
		$\phi$ & Value function params. \\
		$\beta_k$ & Learning rate at step $k$ \\
		$\mathcal{T}$ & Soft Bellman operator \\
		$\mathcal{H}(\pi(\cdot|s))$ & Policy entropy \\
		$\rho_{\pi}$ & State-action distribution \\
		$O$ & Complexity upper bound \\
		\bottomrule
	\end{tabular}
	\label{table:symbol_notation}
\end{table}

\section{Experiments}
\label{sec:experiment}

We conducted a systematic evaluation and validation of MOSEAC’s performance on
task completion ability (e.g., navigating to target locations, avoiding yellow
lines), resource consumption (measured by the number of steps required to
complete the task), and time cost. Additionally, we assessed the trajectory and
control output similarity to report the differences between simulation and
reality. Finally, we compared the computing resource usage between variable and
fixed RL models.

\begin{figure*}[!t]
	\centering
	\includegraphics[width=6.5in]{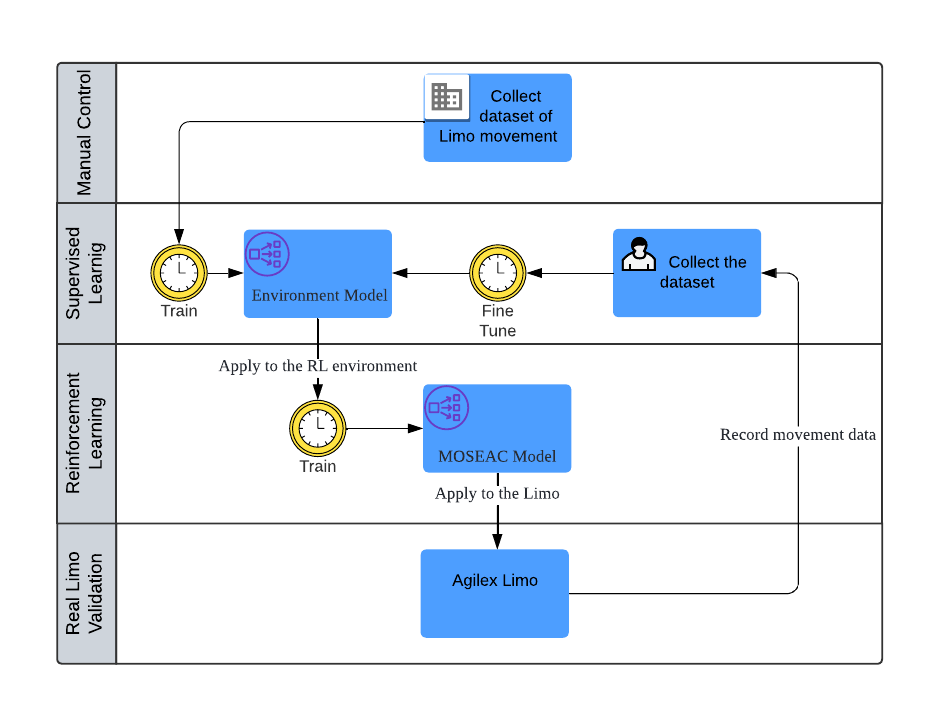}
	\caption{The workflow for our MOSEAC implementaion to the Agilex Limo, we 
		use a joystick to control the Limo movement for the initial 
		environmental data collection (top).}
	\label{fig:workflow}
\end{figure*}

\autoref{fig:workflow} illustrates our workflow. Our simulation to reality
approach involves a process to ensure the reliability 
of MOSEAC when applied to a real-world Limo. The workflow includes
the following steps:

\begin{enumerate}
\item \textbf{Manual Control and Data Collection}: Initially, we manually
  control the Limo to collect a dataset of its movements.
\item \textbf{Supervised Learning}: This dataset is then used to train an
  environment model in a supervised learning manner.
\item \textbf{Reinforcement Learning}: The trained environment model is applied
  in the reinforcement learning environment to train the MOSEAC model.
\item \textbf{Validation and Fine-Tuning}: The MOSEAC model is applied to the
  real Limo. Fine-tuning is performed based on the recorded movement data to
  ensure accurate translation from simulation to real-world application.
\item \textbf{Application}: Finally, the refined MOSEAC model is deployed for
  practical use and validated through real-world tasks.
\end{enumerate}

\subsection{Environment Setup}
\label{subsec:real}

\begin{figure}[!t]
	\centering
	\includegraphics[width=3.0in]{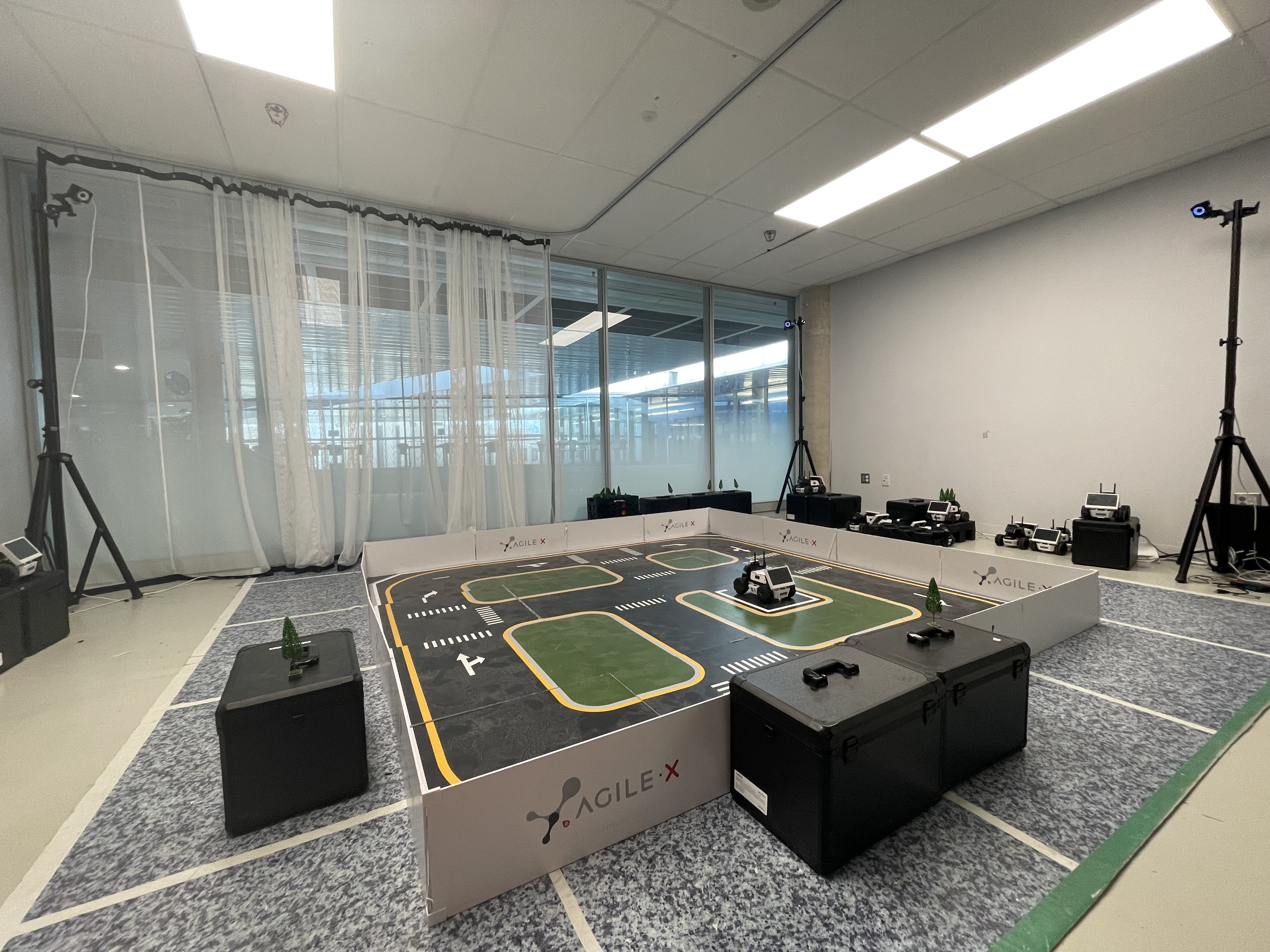}
	\caption{This photo depicts the real-world environment used to validate the 
		performance of MOSEAC on the Agilex Limo. The cameras on the left, 
		right, and middle stands are three of the four cameras comprising the 
		OptiTrack positioning system.}
	\label{fig:map}
\end{figure}

\autoref{fig:map} shows Limo's real-world validation environment. We used an
OptiTrack \cite{optitrack} system for real-time positioning.
We use a 3x3 meter global coordinate frame with its center aligned with
the center of the 2D map, recording the positions of yellow lines within the
map. The Limo's navigable area is confined within the map boundaries, excluding
four enclosed regions. Specific values of these positions, the specifications of
the Agilex Limo, and other metrics can be found in Appendix \ref{appendix:real}.

\subsection{Simulation Environment}
\label{subsec:sim}

For our kinematic modeling, we use an Ackerman model:
\begin{definition}
	\label{def:ackerman_model}	
	The Ackerman kinematic model.
	
	\autoref{tab:symbols} gives the symbols and descriptions.
	\begin{table}[h!]
		\centering
		\caption{List of Symbols and Descriptions}
		\resizebox{\columnwidth}{!}{
			\begin{tabular}{|c|l|}
				\hline
				\textbf{Symbol} & \textbf{Description} \\
				\hline
				$X$ and $Y$ & Current positions of Limo \\
				\hline
				$V$ & Current velocity of Limo \\
				\hline
				$\theta$ & Current angular velocity of Limo \\
				\hline
				$\Delta t$ & Control duration \\
				\hline
				$V_{target}$ & Control linear velocity \\
				\hline
				$\delta$ & Control angular velocity \\
				\hline
				$\mu_k$ & Coefficient of kinetic friction \\
				\hline
				$P$ & Power factor (which can be negative) of Limo \\
				\hline
				$g = 9.81 \, \text{m/s}^2$ & Gravitational acceleration \\
				\hline
				$M = 4.2 \, \text{kg}$ & Mass of the Limo (measured) \\
				\hline
				$L = 0.204 \, \text{m}$ & Distance between the centers of the front  \\
				                        & and rear wheels (measured)  \\
				\hline
			\end{tabular}
		}
		\label{tab:symbols}
	\end{table}
	
	The forces and accelerations are calculated as:
	\begin{equation}
		\begin{aligned}
			F_{friction} &= -\mu_k M g \\
			a_{friction} &= \frac{F_{friction}}{M} = -\mu_k g \\
			a_{power} &= \frac{P}{M} \\
			a_{net} &= a_{power} + \frac{V_{target} - V}{\Delta t} + a_{friction}
		\end{aligned}
	\end{equation}
	
	The updates for velocity and position are:
	\begin{equation}
		\begin{aligned}
			V_{new} &= V + a_{net} \Delta t \\
			X_{new} &= X + V_{new} \cos(\theta_{new}) \Delta t \\
			Y_{new} &= Y + V_{new} \sin(\theta_{new}) \Delta t
		\end{aligned}
	\end{equation}
	
	This formulation accounts for the dynamics of the Limo vehicle, considering
  friction, power, and vehicle mass, ultimately predicting the new
  position. This kinematic model is referred tp as $M_{Ackerman}$.
	
\end{definition}

We employ supervised learning \cite{rybczak2024survey, imece2022} to model the
motion dynamics from the real-world environment to the simulation environment.
This model predicts kinematic and physical information, $\mu_k$ and $P$, across
different regions within the environment. We select a Transformer Model
\footnote{Our code is publicly available on
  \href{https://github.com/alpaficia/MOSEAC_Limo}{GitHub} with the shape and
  related hyperparameters of this Transformer Model inside.}
\cite{vaswani2023attention} to predict these data. The input, output, and its
loss function are defined in Definition \autoref{def:transformer}.

\begin{definition}
	\label{def:transformer}
	
	The input $\mathbf{X}$ is:
		
		\[
		\mathbf{X} = \begin{pmatrix} X, Y, V, \theta, \Delta t, V_{target}, 
		\delta \end{pmatrix}
		\]
	
	The output $\mathbf{O}$ is:
		
		\[
		\mathbf{Y} = \begin{pmatrix} \mu_k, P \end{pmatrix}
		\]
	
	Using $M_{Ackerman}$ from Definition \autoref{def:ackerman_model}, the loss 
	function of the Transformer model $T$ is computed as:
	
	\begin{equation}
		\begin{aligned} 
			\text{predict} &= M_{Ackerman}(\mathbf{X}, \mathbf{Y}, g) \\
			\text{loss} &= \frac{1}{N} \sum_{i=1}^{N} \left( \text{predict}_i - 
			\text{target}_i \right)^2
		\end{aligned}
	\end{equation}
	
	The loss function is defined as the Mean Squared Error (MSE) between the
  predicted positions and the target positions recorded in the dataset:
	
	where:
	\begin{itemize}
  \item $\text{predict}_i$ are the predicted positions for the $i$-th data point
  \item $\text{target}_i$ are the actual positions for the $i$-th data point
	\end{itemize}
	
\end{definition}

After learning the kinematic model, we need to address the state definition of
MOSEAC. It should include the necessary environmental information to enable the
model to understand the positions of obstacles in the environment. Although we
have recorded all the positions of these yellow lines, it is not reasonable to
input them directly as state values. This would compromise the generality of our
approach, and a large amount of invariant fixed position information might cause
the neural network processing units to struggle with capturing the critical
information steps or lead to overfitting \cite{impact_distribution_shift_2024,
  katakura2023rl_dynamic_state_space}.

\begin{figure}[!t]
	\centering
	\includegraphics[width=3.5in]{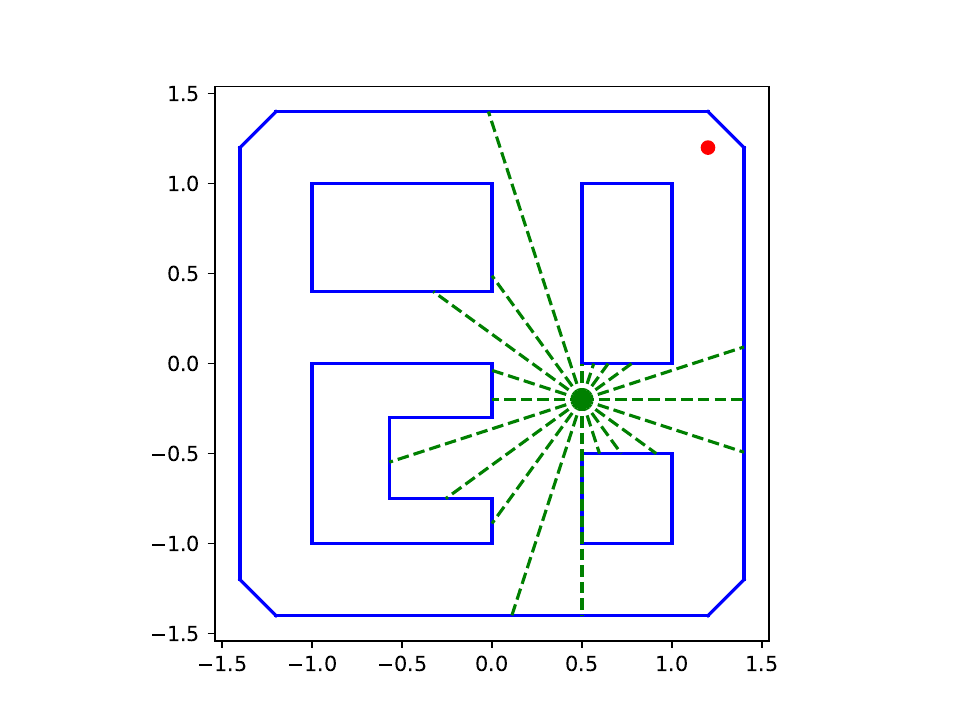}
	\caption{The simulated lidar system generates 20 rays from Limo's position,
    calculates their intersections with environment enclosed regions, and
    returns the nearest intersection points for each ray.}
	\label{fig_radar}
\end{figure}

We developed a simulated lidar system centered on Limo's current position. This
system generates 20 rays, similar to lidar operation, to detect intersections
with enclosed regions in the environment. These intersections provide
information about restricted areas, as shown in \autoref{fig_radar}.

Since the enclosed regions data in our simulation are based on real-world
measurements, the gap between the virtual and real world is negligible. We
designated eight turning points on this map as navigation endpoints, with a
specific starting point for Limo.

The state dimensions for our MOSEAC model include:
\begin{inparaitem}
	\item robot current position,
	\item goal position (chosen among 8 random locations),
	\item linear velocity,
	\item steering angle,
	\item 20 lidar points, and
	\item previous control duration, linear and angular velocity.
\end{inparaitem}
The action dimensions include control duration, linear velocity, and angular
velocity.

Let $R_t$ represent the task reward and $T$ denote the termination flag (where 1
indicates termination and 0 indicates continuation). The function
$d(\mathbf{p}_1, \mathbf{p}_2)$ measures the distance between positions
$\mathbf{p}_1$ and $\mathbf{p}_2$. The variables $\mathbf{a}_{\text{new}}$ and
$\mathbf{t}$ denote the new position of the agent and the target position,
respectively. The initial distance from the starting position to the target
position is $d_0$, and $\delta$ specifies the threshold distance determining
whether Limo is sufficiently close to a point. Additionally, $C_{\text{inner}}$
indicates a collision with enclosed regions, while $C_{\text{outer}}$ indicates
a collision with the map boundary.

Implementing a penalty mechanism instead of a termination mechanism in our
reward definition offers significant advantages: the penalty mechanism provides
incremental feedback, allowing the model to continue learning even after errors,
thereby avoiding frequent resets that could disrupt the training process. This
approach reduces overall training time and encourages the agent to explore a
broader range of strategies, leading to a more comprehensive understanding of
the environment's dynamics. However, we still terminate an experiment if the
robot wonders outside the map boundary to stop meaningless exploration.
The reward $R_t$ is:
\begin{equation}
R_t = 
	\begin{cases} 
		R_{\text{boundary\_penalty}}, & \text{if } C_{\text{inner}}(\mathbf{a}_{\text{new}}) \\
		R_{\text{success\_reward}}, & \text{if } d(\mathbf{a}_{\text{new}}, \mathbf{t}) \leq \delta \\
		R_{\text{failure\_penalty}}, & \text{if } C_{\text{outer}}(\mathbf{a}_{\text{new}}) \\
		R_{\text{distance\_reward}} = d_0 - d(\mathbf{a}_{\text{new}}, \mathbf{t}), & \text{otherwise}
	\end{cases}
\end{equation}
The termination flag $T$ is determined as:
\begin{equation}
T = 
	\begin{cases} 
		0, & \text{if } C_{\text{inner}}(\mathbf{a}_{\text{new}}) \\
		1, & \text{if } d(\mathbf{a}_{\text{new}}, \mathbf{t}) \leq \delta \\
		1, & \text{if } C_{\text{outer}}(\mathbf{a}_{\text{new}}) \\
		0, & \text{otherwise}
	\end{cases}
\end{equation}

The initial data collected using remote control was skewed for short action 
durations, causing some discrepancies between the physical robot behaviour and 
the behaviour in simulation. To address these discrepancies we fine-tuned a 
Transformer model that approximates Limo's kinematic model by collecting new 
real-world movement data.

We begin by freezing all the layers of the Transformer model except the final
fully connected layer. This ensures that the initial generalized features are
preserved while adapting the model's outputs to the specific characteristics of
the Limo dataset. The training process is:
\begin{compactenum}
	\item \textbf{Freeze Initial Layers}: Initially, all layers except the final 
	fully connected layer are frozen. $\theta_i$ represents the 
	parameters of the $i$-th layer:
	\begin{equation}
	\text{if } i < N_{\text{unfrozen}}, \quad \nabla_{\theta_i} L = 0
	\end{equation}
	where $L$ is the loss function and $N_{\text{unfrozen}}$ is the number of 
	layers that are not frozen.
	
\item \textbf{Gradual Unfreezing}: After training the final layer, we
  incrementally unfreeze the preceding layers and fine-tune the model further.
  This is done iteratively, where at each stage, one additional layer is
  unfrozen and the model is re-trained. The number of unfrozen layers increases
  until all layers are eventually fine-tuned:
	\begin{equation}
	\text{if } i \geq N_{\text{unfrozen}}, \quad \nabla_{\theta_i} L \neq 0
	\end{equation}
	
	\item \textbf{Early Stopping}: To prevent overfitting, we employ early 
	stopping. If the validation loss does not improve for a specified number of 
	epochs, training is halted:
	\begin{equation}
	\text{if } \Delta L_{\text{valid}} > 0 \text{ for } n \text{ epochs}, 
	\quad \text{stop training}
	\end{equation}
	where $\Delta L_{\text{valid}}$ is the change in validation loss and $n$ is 
	the patience parameter.
	
	\item \textbf{Evaluation}: The model's performance is evaluated on a test 
	dataset after each training phase. The final model is selected based on the 
	lowest validation loss.
\end{compactenum}	

The total training process can be described by the following optimization
problem:
\begin{equation}
	\min_{\theta} \sum_{(x,y) \in \mathcal{D}_{\text{train}}} \mathcal{L}
	(f_{\theta}(x), y)
\end{equation}
where $\mathcal{D}_{\text{train}}$ is the training dataset, $f_{\theta}$ is the
model with parameters $\theta$, and $\mathcal{L}$ is the loss function
\ref{def:transformer}. The fine-tuning process specifically involves updating
the parameters layer by layer, ensuring that the generalized pre-trained
knowledge is gradually adapted to the specific nuances of the new Limo dataset.
	
This fine-tuning strategy allows the use of pre-trained models, adapting them to
new tasks with potentially smaller datasets while maintaining robust performance
and preventing overfitting.

For more detailed information
please refer to Appendix~\ref{appendix:sim}.

\section{Results}
\label{sec:result}

We conducted six experiments involving MOSEAC, MOSEAC without the $\alpha_{max}$
limitation, SEAC, CTCO, and SAC (at 20 Hz and 60 Hz, SAC20 and SAC60,
respectively) within the simulation environment and applied these trained models
to the real Limo. These training tests were performed on a computer equipped
with an Intel Core i5-13600K CPU and an Nvidia RTX 4070 GPU running Ubuntu 22.04
LTS. The deployment tests were conducted on an Agilex Limo equipped with a
Jetson Nano \cite{nvidia_jetson_nano} (with a Cortex-A57 CPU and Maxwell 128
cores GPU) running Ubuntu 18.04 LTS.

We trained our RL model using PyTorch \cite{paszke2019pytorch} within a
Gymnasium environment \cite{towers_gymnasium_2023}. Subsequently, we converted
the model parameters to ONNX format \cite{paszke2019pytorch}. Finally, we
utilized TensorRT \cite{tensorrt} to build an inference engine from the ONNX
model in the AgileX Limo local environment and deployed it for RL navigation
tasks \footnote{Our code is publicly available on
  \href{https://github.com/alpaficia/MOSEAC_Limo}{GitHub} with the
  hyperparameters of the MOSEAC model inside.}. More system information details
can be found in Appendix \ref{appendix:os}.

\begin{figure}[!t]
	\centering
	\includegraphics[width=3.5in]{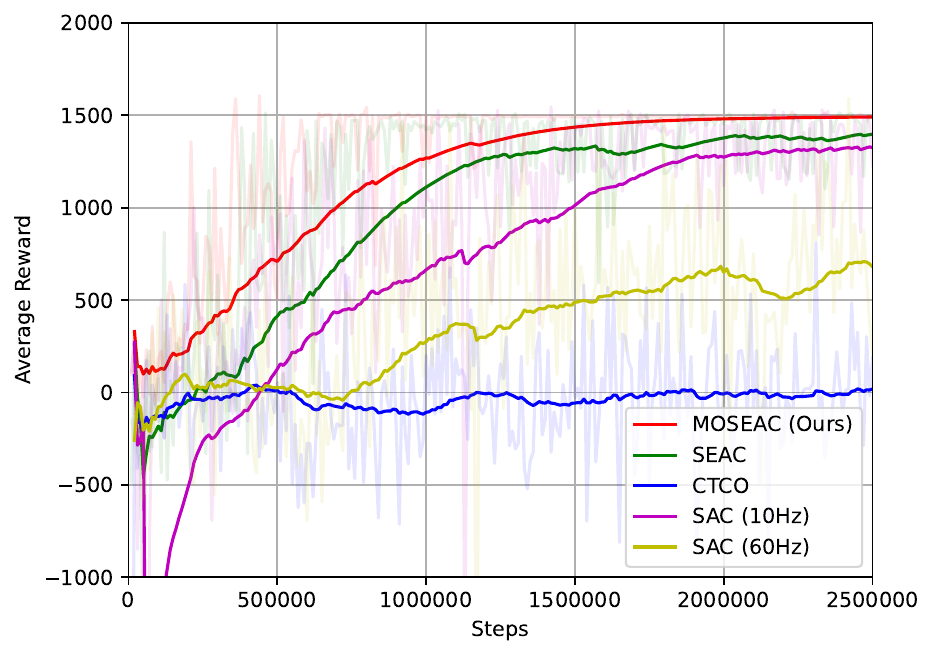}
	\caption{Average returns of 5 reinforcement learning algorithms over 2.5M 
		steps during training.}
	\label{fig:limo_return}
\end{figure}

\begin{figure}[!t]
	\centering
	\includegraphics[width=3.5in]{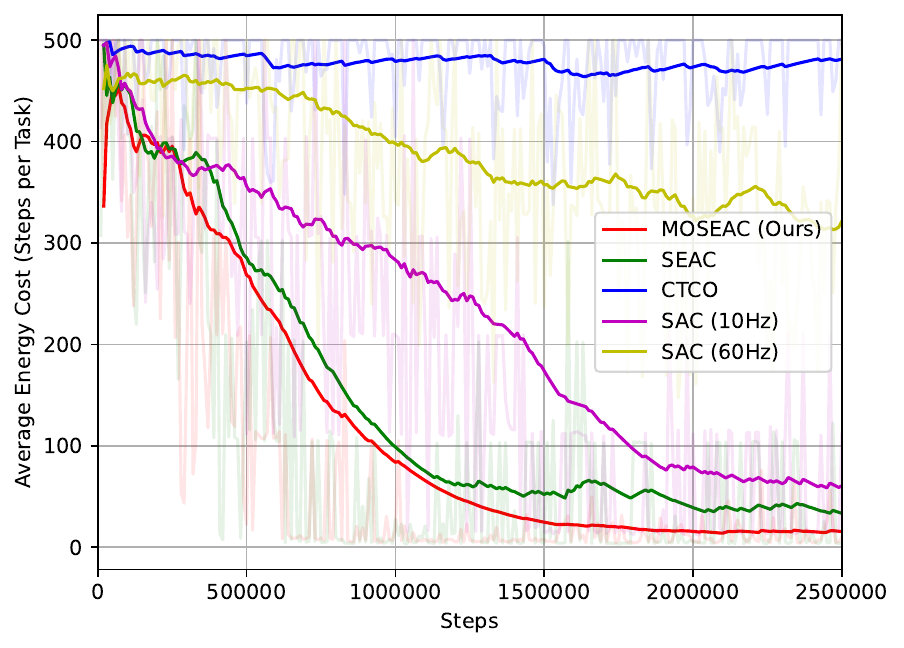}
	\caption{Average energy costs of 5 reinforcement learning algorithms over 
		2.5M steps during training.}
	\label{fig:limo_cost}
\end{figure}

\autoref{fig:limo_return} and \autoref{fig:limo_cost} illustrate the results of 
the training process. Several key insights can be drawn from these figures:

\autoref{fig:limo_return} indicates that MOSEAC consistently improves over time
compared to the other algorithms, suggesting that MOSEAC adapts well to the
environment and maintains a stable learning curve. While SEAC also shows stable
performance, it converges slightly slower than MOSEAC. Additionally,
\autoref{fig:limo_return} illustrates that MOSEAC exhibits higher action
duration robustness than CTCO. MOSEAC utilizes a fixed discount factor
($\gamma$), ensuring stable long-term planning capabilities without the negative
impact of varying $\gamma$. In contrast, CTCO's performance is susceptible to
the choice of action duration range, significantly affecting its $\gamma$. As
training progresses, CTCO tends to favor smaller $\gamma$ values, placing
greater demands on its $\tau$ parameter that controls the range of $\gamma$.
This sensitivity makes CTCO less adaptable to diverse environments, whereas
MOSEAC maintains consistent performance. Furthermore, \autoref{fig:limo_cost}
provides a comparative analysis of energy costs among the different algorithms.
MOSEAC demonstrates lower energy costs per task, indicating higher energy
efficiency during training. This efficiency is critical for practical
applications where compute resource consumption is a concern.

\begin{figure}[!t]
	\centering
	\includegraphics[width=3.5in]{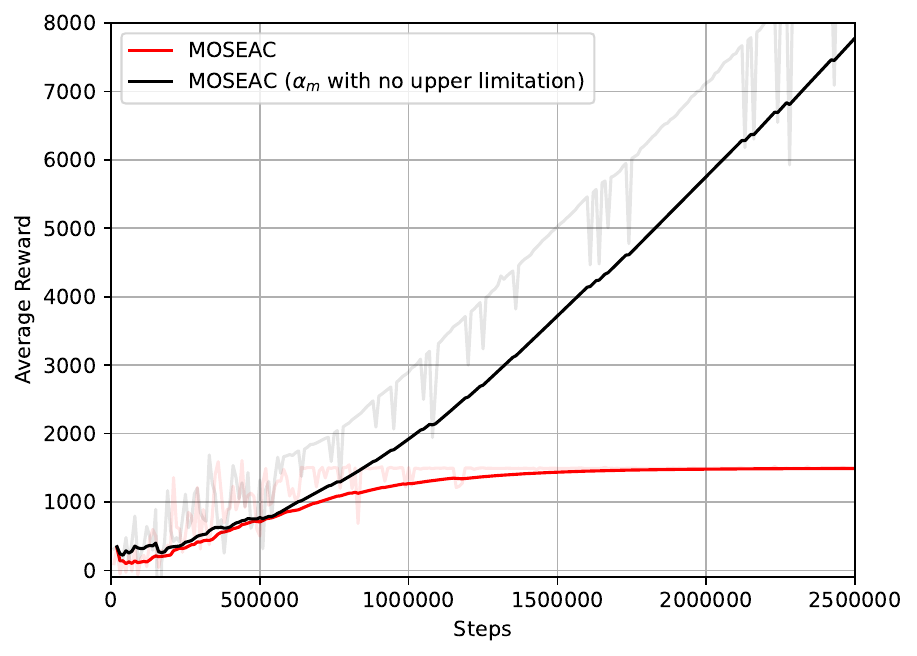}
	\caption{Average returns of MOSEAC and MOSEAC (without $\alpha_m$ 
		limitation) in 2.5M steps during the training.}
	\label{fig:compare_return}
\end{figure}

We compared the effects of imposing an upper limit on the parameter $\alpha_m$
versus allowing it to increase without restriction in the context of our MOSEAC
algorithm. Our findings indicate significant differences in performance
stability and energy efficiency between the two approaches.

Introducing an upper limit on $\alpha_m$ (denoted as $\alpha_{max}$) is required
to prevent reward explosion. \autoref{fig:compare_return} shows that MOSEAC has
consistent and stable improvement in average reward. In contrast, MOSEAC without
the upper limit initially followed a similar trend but eventually diverged,
leading to instability and potential reward explosion. This divergence suggests
that without the upper limit, $\alpha_m$ may increase uncontrollably,
destabilizing the reward structure.

\begin{figure}[!t]
	\centering
	\includegraphics[width=3.5in]{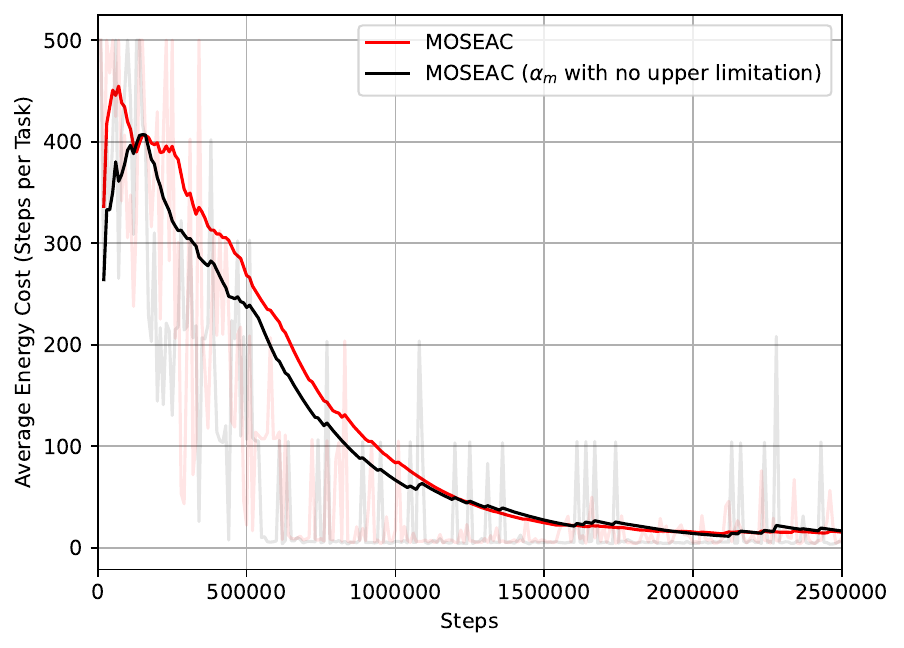}
	\caption{Average Energy cost of MOSEAC and MOSEAC (without $\alpha_m$ 
		limitation) in 2.5M steps during the training.}
	\label{fig:compare_cost}
\end{figure}

However, as shown in Figure~\ref{fig:compare_cost}, MOSEAC without an upper
limit on $\alpha_m$ generally exhibited lower average energy costs compared to
MOSEAC with the upper limit. This suggests that while the absence of an upper
limit on $\alpha_m$ may lead to reward explosion, it does not significantly
impair task performance in practice.

We update $\alpha_m$ based on the declining reward trend. However, in random
multi-objective tasks with varying target locations, the average reward can
fluctuate significantly, causing $\alpha_m$ to be updated continuously. This can
lead to an unbounded increase in reward during training when $\alpha_m$ is
unrestricted, amplifying the reward signal. The increased gradient variability
results in rapid strategy adjustments, enhancing short-term energy efficiency.
Despite the potential for reward explosion, the impact on task performance is
minimal, with notable improvements in energy efficiency.

\begin{figure}[!t]
	\centering
	\includegraphics[width=3.5in]{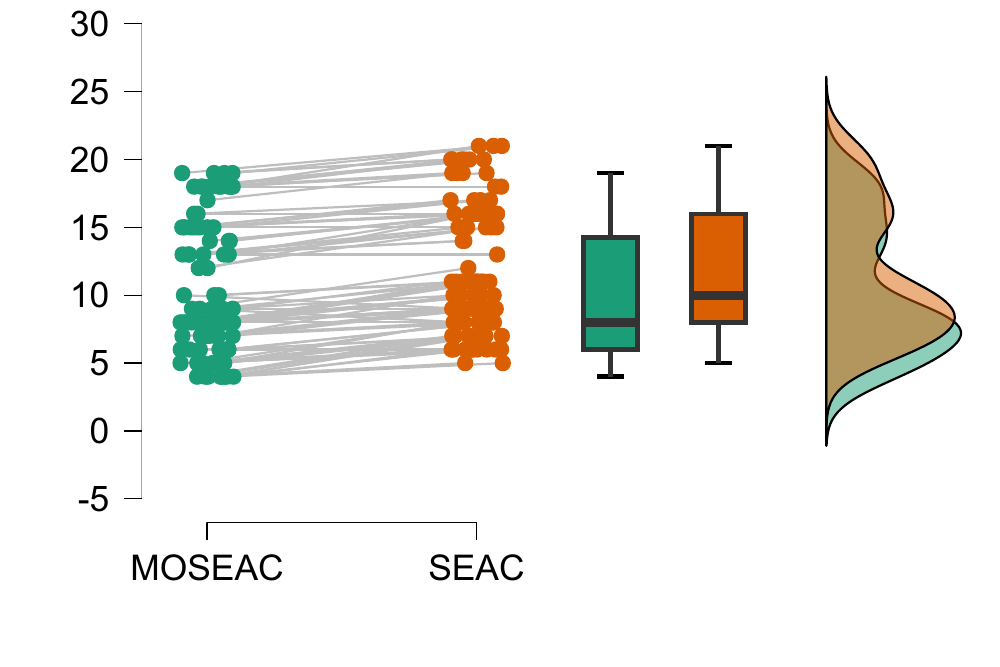}
	\caption{Energy cost (as number of time steps) for 100 random tasks. We use 
		the same seed for MOSEAC and SEAC to ensure that the tasks are the same 
		for the two algorithms.}
	\label{fig:limo_energy}
\end{figure}

\begin{figure}[!t]
	\centering
	\includegraphics[width=3.5in]{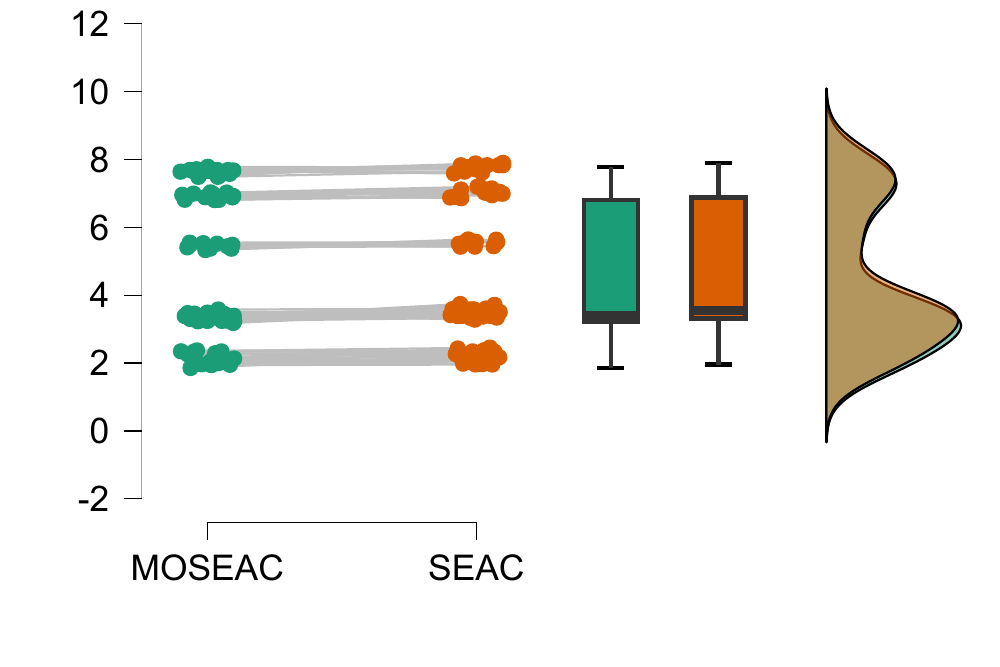}
	\caption{Time cost for 100 random tasks. We use the same seed for MOSEAC 
		and SEAC to ensure that the tasks are the same for the two algorithms.}
	\label{fig:limo_time}
\end{figure}

We compared MOSEAC's task performance with the best-performing SEAC after
training. Due to the poor performance of CTCO and SAC, we excluded them from the
analysis.

\autoref{fig:limo_energy} and \autoref{fig:limo_time} illustrate the energy and
time distributions for MOSEAC and SEAC methods. Given the lack of normality in
the data distribution, we use the Wilcoxon signed-rank test to compare the
paired samples. For energy consumption (\autoref{fig:limo_energy}), MOSEAC
demonstrates significantly lower median and overall energy usage than SEAC
($W=35.0,z=-7.904,p<.001$). For time efficiency (\autoref{fig:limo_time}),
MOSEAC also shows a significantly lower median time, indicating quicker task
completion than SEAC ($W=502.0,z=-6.956,p<.001$). The descriptive statistics and
test results are summarized in Appendix \ref{appendix:statistic_result}.

 The improved performance of MOSEAC over SEAC can be attributed to its reward
function. While SEAC’s reward function is linear, combining task reward, energy
penalty, and time penalty independently, MOSEAC introduces a multiplicative
relationship between task reward and time-related reward. This non-linear
interaction enhances the reward signal, particularly when both task performance
and time efficiency are high, and naturally balances these factors. By keeping
the energy penalty separate, MOSEAC maintains flexibility in tuning without
complicating the relationship between time and task rewards. This design allows
MOSEAC to more effectively guide the agent’s decisions, resulting in better
energy efficiency and task completion speed in practical applications.

We conducted real-world tests on the Agilex Limo using ROS~\cite{quigley2009ros}\footnote{Our code for ROS workspace is publicly available
  on \href{https://github.com/alpaficia/MOSEAC_Limo}{GitHub}, which also
  provides support for ROS 2 \cite{ros2} and Docker \cite{docker}}. The Limo
robot navigated to random and distinct endpoints using the MOSEAC model as its
control policy for Ackerman steering~\footnote{Video avaliable here:
  \url{https://youtu.be/VhTa66WqxoU}}. We collected these data and calculated
trajectory and control output similarities with respect to simulation. 

For trajectory similarity, we comine all trajectory data in to one trajetory, 
the ATE (Average Trajectory Error)\cite{ryan2004review} is
$0.036$ meters, indicating minimal deviations between the actual and
simulated paths. Additionally, the Dynamic Time Warping (DTW) 
\cite{muller2007dynamic} value of $0.531$ supports the high degree of 
similarity between the temporal sequences of the trajectories. These metrics 
suggest that our method enables the Limo to follow the planned paths with high 
precision, closely mirroring the simulation.

Regarding control output similarity, the Mean Absolute Error (MAE) is $0.002$,
and the Mean Squared Error (MSE) is $4.49E-05$. These low error values indicate
that the control outputs in the real world closely match those in the
simulation. Our method effectively minimizes the discrepancies in control
signals, ensuring that the robot's actions are consistent across different
environments.

%


Overall, the empirical data supports the theoretical claims regarding MOSEAC's
performance in trajectory fidelity and control output consistency.

\begin{figure}[!t]
	\centering
	\includegraphics[width=3.5in]{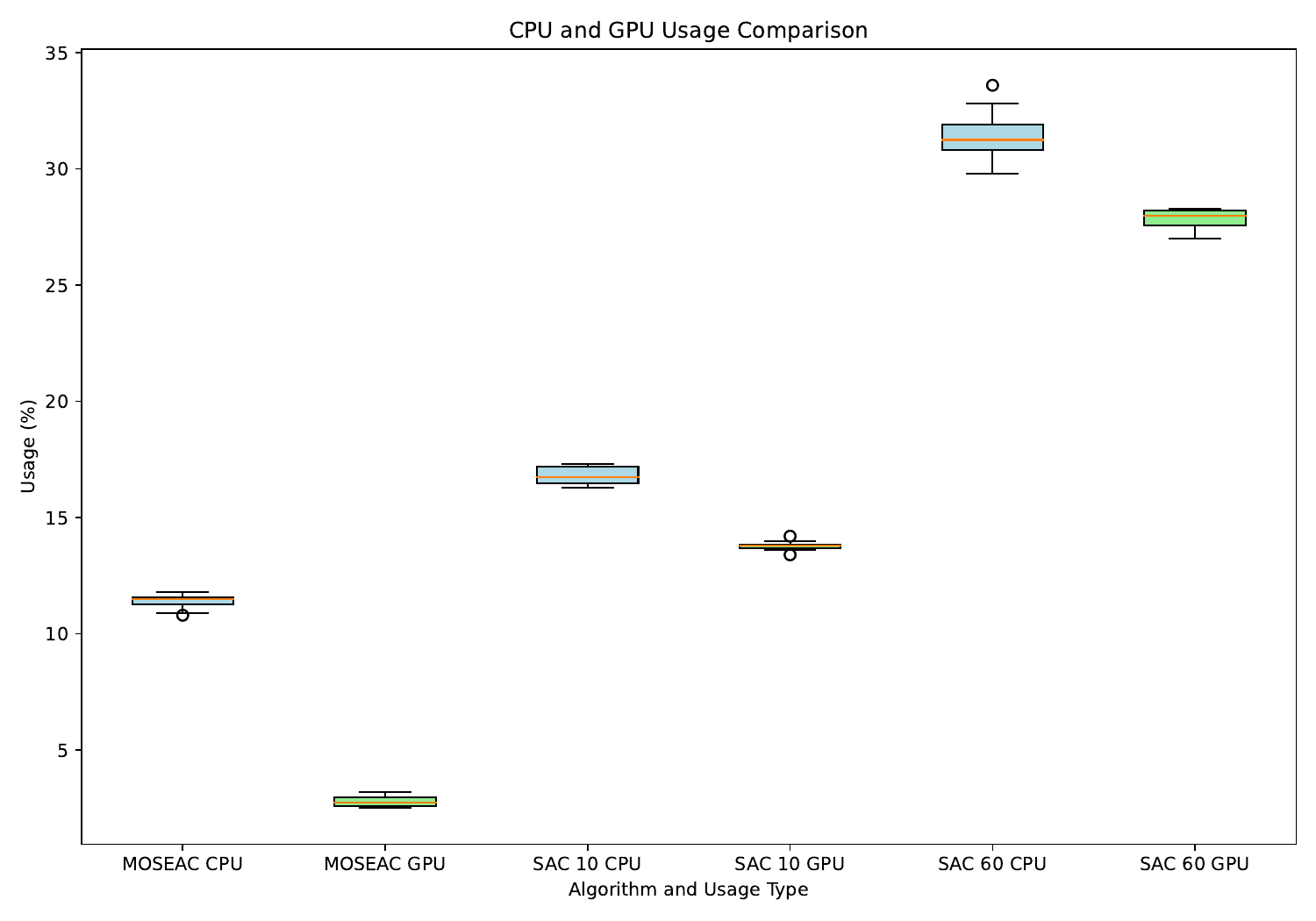}
	\caption{Comparison of Average Compute Resource Usage Across Different Methods}
	\label{fig:compute_resources_usage}
\end{figure}

We also recorded the computing resource usage, highlighting the efficiency of
the MOSEAC algorithm in terms of energy consumption, particularly computational
energy. \autoref{fig:compute_resources_usage} provides a comparison of the
average usage per second of CPU and GPU resources between MOSEAC and SAC (Soft
Actor-Critic) algorithms running at different frequencies (10~Hz and 60~Hz).


The CPU usage data reveals that MOSEAC significantly reduces computational load
compared to SAC at both 10 Hz and 60 Hz. Specifically, MOSEAC utilizes only
$11.40\% \pm 0.12$ of the CPU resources, whereas SAC requires $16.80\% \pm 0.14$
at 10 Hz and $31.41\% \pm 1.47$ at 60 Hz. Similarly, the GPU usage data
indicates that MOSEAC is more efficient, using only $2.80\% \pm 0.07$ compared
to SAC's $13.79\% \pm 0.05$ at 10 Hz and $27.86\% \pm 0.19$ at 60 Hz. This
reduction conserves energy (increasing battery life) and frees processing power
for tasks like perception and communication. The descriptive statistics and
test results are summarized in Appendix \ref{appendix:compute_statistic}.




\section{Conclusions}
\label{sec:conclusions}

In this paper, we presented the Multi-Objective Soft Elastic Actor-Critic
(MOSEAC) algorithm, a Variable Time Step Reinforcement Learning (VTS-RL) method
designed with adaptive hyperparameters that respond to observed reward trends
during training. Our analysis included theoretical performance
guarantees, convergence analysis, and practical validation through simulations
and real-world navigation tasks using a small rover.

We compared MOSEAC with other VTS-RL algorithms, such as SEAC 
\cite{wang2024deployable} and CTCO \cite{karimi2023dynamic}. While SEAC and 
CTCO improved over traditional fixed-time step methods, they still required 
extensive hyperparameter tuning and did not achieve the same level of 
efficiency and robustness as MOSEAC. SEAC's reward structure, though effective, 
was less adaptable, and CTCO's sensitivity to action duration further limited 
its practical application.

Additionally, the empirical data demonstrated that MOSEAC significantly
outperforms traditional SAC algorithms, particularly in terms of computational
resource efficiency. MOSEAC's reduced CPU and GPU usage frees up resources for
other critical tasks, such as environment perception and map reconstruction,
enhancing the robot's operational efficiency and extending battery life. This
makes MOSEAC highly suitable for long-term and complex missions.

Our findings validate the robustness and applicability of MOSEAC in real-world
scenarios, providing strong evidence of its potential to lower hardware
requirements and improve data efficiency in reinforcement learning deployments.

Our future work will further refine the algorithm, particularly in the adaptive
tuning of hyperparameters, to enhance its performance and applicability. We aim
to apply MOSEAC to a broader range of robotic projects, including smart cars and
robotic arms, to fully leverage its benefits in diverse practical settings.

\section{Acknowledgement}
The authors sincerely thank Yann Bouteiller, Guillaume Ricard, and Wenqiang Du
for their invaluable assistance with the sim-to-real method discussions, AgileX
Limo usage, and ROS support. Their expertise and dedication were crucial in
achieving high-quality sim-to-real deployment, which was essential for the
success of this research. We greatly appreciate their significant time and
effort invested in this project.

\appendices
\section{Analysis of the Adaptive $\alpha_m$ Adjustment in MOSEAC}
\label{appendix:adaptive}

In MOSEAC we use an adaptive adjustment scheme for $ \alpha_m $, specifically a
simple linear increment with an upper limit $ \alpha_{\max} $. The inclusion of
$ \psi $ and the reward adjustment strategy inherently involves Pareto
optimization. Below, we provide a mathematical analysis of the stability of our
scheme and its relation to the Pareto front.

\subsection{Adaptive Adjustment Scheme for $ \alpha_m $}

Let $ \alpha_m $ be adjusted linearly over time with an upper limit $
\alpha_{\max} $:
\begin{equation}
\alpha_m(t) = \min(\alpha_{m,0} + kt, \alpha_{\max})
\end{equation}
where $ \alpha_{m,0} $ is the initial value, $ k $ is the increment rate, and $
t $ represents the time or iteration index.

The stability of this linear adjustment can be analyzed by examining the impact
on the reward function $ R $ and the policy updates.

\subsection{Reward Function and $ \psi $}

The reward function $ R $ in the MOSEAC algorithm can be expressed as:
\begin{equation}
R = r + \psi
\end{equation}
where $ r $ is the immediate reward and $ \psi $ is an adjustment term that 
influences the reward based on the adaptive $ \alpha_m $.

Given the linear increment of $ \alpha_m $, the adjustment term $ \psi $ can be 
represented as:
\begin{equation}
\psi(t) = f(\alpha_m(t)) = f(\alpha_{m,0} + kt) \text{ for } \alpha_m(t) 
< \alpha_{\max}
\end{equation}

\begin{equation}
\psi(t) = f(\alpha_{\max}) \text{ for } \alpha_m(t) \geq \alpha_{\max}
\end{equation}
where $ f $ is a function that defines how $ \psi $ depends on $ \alpha_m $.

\subsection{Stability Analysis}

To analyze the stability of the reward function under this adaptive scheme, we 
examine the boundedness and convergence of $ R $. The stability is ensured if 
the cumulative reward remains bounded and the policy converges to an optimal 
policy over time.

\begin{enumerate}
	\item \textbf{Boundedness}: The linear increment of $ \alpha_m $ with an 
	upper limit should ensure that $ R $ does not grow unbounded. This requires 
	that $ f(\alpha_m) $ grows at a controlled rate.
	\begin{equation}
	|R| = |r + f(\alpha_{m,0} + kt)| \leq M \text{ for } \alpha_m(t) < 
	\alpha_{\max}
	\end{equation}
	\begin{equation}
	|R| = |r + f(\alpha_{\max})| \leq M \text{ for } \alpha_m(t) \geq 
	\alpha_{\max}
	\end{equation}
	where $ M $ is a constant, indicating that $ R $ remains bounded.
	\item \textbf{Convergence}: The policy $ \pi $ should converge to an optimal
	policy $ \pi^* $. The convergence is influenced by the adjustment term 
	$ \psi $ and the learning rate $ \alpha_m $.
	\begin{equation}
	\lim_{t \to \infty} \pi(t) = \pi^*
	\end{equation}
	If $ \alpha_m $ is adjusted linearly with an upper limit, the learning rate 
	should decrease over time to ensure convergence.
\end{enumerate}

\subsection{Multi-Objective Optimization}

The multi-objective optimization aspect of the MOSEAC algorithm arises from the
trade-off between multiple objectives in the reward function. The inclusion of
$\psi$ introduces a multi-objective optimization problem, where the algorithm
aims to optimize the cumulative reward while balancing different aspects
influenced by $\psi$.

The Pareto front represents the set of optimal policies where no single
objective can be improved without degrading another~\cite{pareto1971manual}. The
linear increment of $ \alpha_m $ with an upper limit and the presence of $ \psi
$ ensure that the algorithm explores the policy space effectively, converging to
solutions on the Pareto front~\cite{deb2002fast}.


\section{Real Environment}
\label{appendix:real}

\autoref{tab:zone} displays the coordinate information of the enclosed regions 
in the real environment; \autoref{tab:limo_specs} shows the specifications of 
the Agilex Limo; \autoref{tab:optitrack_metrics} shows the key performance 
metrics of the OptiTrack system.

\begin{table}[h!]
	\centering
	\caption{Enclosed regions information (yellow lines arrays) in the real environment (in meters)}
	\resizebox{\columnwidth}{!}{
		\begin{tabular}{>{\raggedright\arraybackslash}p{4cm} >{\centering\arraybackslash}p{7cm}}
			\toprule
			\textbf{Zone} & \textbf{Coordinates} \\
			\midrule
			zone\_left\_down & [[0.0, 0.0], [-1.0, 0.0]], [[-1.0, 0.0], [-1.0, -1.0]], [[-1.0, -1.0], [0.0, -1.0]], [[0.0, -1.0], [0.0, -0.75]], [[0.0, -0.75], [-0.57, -0.75]], [[-0.57, -0.75], [-0.57, -0.3]], [[-0.57, -0.3], [0.0, -0.3]], [[0.0, -0.3], [0.0, 0.0]] \\
			zone\_left\_up & [[0.0, 0.4], [0.0, 1.0]], [[0.0, 1.0], [-1.0, 1.0]], [[-1.0, 1.0], [-1.0, 0.4]], [[-1.0, 0.4], [0.0, 0.4]] \\
			zone\_right\_up & [[0.5, 0.0], [1.0, 0.0]], [[1.0, 0.0], [1.0, 1.0]], [[1.0, 1.0], [0.5, 1.0]], [[0.5, 1.0], [0.5, 0.0]] \\
			zone\_right\_down & [[0.5, -1.0], [0.5, -0.5]], [[0.5, -0.5], [1.0, -0.5]], [[1.0, -0.5], [1.0, -1.0]], [[1.0, -1.0], [0.5, -1.0]] \\
			\bottomrule
	\end{tabular}}
	
	\label{tab:zone}
\end{table}

\begin{table}[h!]
	\centering
	\caption{AgileX Limo Specifications \cite{agilex_limo}}
	\resizebox{\columnwidth}{!}{
		\begin{tabular}{|p{4cm}|p{8cm}|}
			\hline
			\textbf{Category} & \textbf{Specifications} \\
			\hline
			Dimensions & 322mm x 220mm x 251mm \\
			\hline
			Weight & 4.2 kg \\
			\hline
			Maximum Speed & 1 m/s \\
			\hline
			Maximum Climbing Capacity & 25° (omni-wheel, differential, Ackermann steering), 40° (tracked mode) \\
			\hline
			Ground Clearance & 24 mm \\
			\hline
			Battery & 12V Li-ion 5600mAh \\
			\hline
			Run Time & 40 minutes of continuous operation \\
			\hline
			Standby Time & 2 hours \\
			\hline
			Charging Time & 2 hours \\
			\hline
			Operating Temperature & -10°C to 40°C \\
			\hline
			IP Rating & IP22 (splash-proof, dust-proof) \\
			\hline
			Sensors & 
			IMU: MPU6050 \newline
			LiDAR: EAI X2L \newline
			Depth Camera: ORBBEC DaBai \\
			\hline
			CPU & ARM64 Quad-Core 1.43GHz (Cortex-A57) \\
			\hline
			GPU & 128-core NVIDIA Maxwell™ @ 921MHz \\
			\hline
			Communication & Wi-Fi, Bluetooth 5.0 \\
			\hline
			Operating System & Ubuntu 18.04, ROS1 Melodic \\
			\hline
	\end{tabular}}
	\label{tab:limo_specs}
\end{table}

\begin{table}[h!]
	\centering
	\caption{Key Performance Metrics of the OptiTrack System \cite{optitrack}}
	\resizebox{\columnwidth}{!}{
		\begin{tabular}{>{\raggedright\arraybackslash}p{4cm} >{\centering\arraybackslash}p{3cm} >{\raggedright\arraybackslash}p{5cm}}
			\toprule
			\textbf{Performance Metric} & \textbf{Value} & \textbf{Description} \\
			\midrule
			Camera Resolution & Up to 4096x2160 & High resolution for detailed tracking \\
			Frame Rate & Up to 360 FPS & Ensures smooth motion capture \\
			Latency & As low as 3ms & Provides real-time feedback \\
			Field of View (FOV) & 56° to 100° & Suitable for various capture needs \\
			Synchronization Precision & \textless 1µs & Multi-camera synchronized capture \\
			Marker Tracking Accuracy & \textless 0.5mm & Precise 3D spatial positioning \\
			Operating Range & Up to 30m & Suitable for large-scale capture environments \\
			Optical Resolution & 12 MP & High-quality image data \\
			Ambient Light Suppression & Strong & Reduces interference from ambient light \\
			Data Interface & Gigabit Ethernet & Fast data transmission \\
			\bottomrule
	\end{tabular}}
	
	\label{tab:optitrack_metrics}
\end{table}

\section{Simulation Environment}
\label{appendix:sim}

The position of the goal is randomized from these eight points in each episode. 
Setting multiple endpoints offers several advantages. It enhances the navigation
policy's generalization capability by training the agent to adapt to various 
goals rather than a single target. This promotes extensive exploration, 
prevents the agent from getting stuck in local optima, and increases robustness
by preparing the agent to handle dynamic or uncertain target positions in 
real-world deployments \cite{riedmiller2018}.

Additionally, it improves data efficiency by allowing the agent to learn from 
diverse experiences in a single training session. Multiple endpoints provide 
more prosperous reward signals, accelerating learning through varied success and
failure contexts. Moreover, it simulates realistic scenarios where multiple 
destinations are shared,  thus increasing the practical value of the trained 
model. Lastly, this approach raises task complexity, challenging the agent to 
develop more sophisticated strategies and ultimately enhancing overall 
performance \cite{bellemare2016}.

\begin{table}[h!]
	\centering
	\caption{List of goal positions with their coordinates}
	\begin{tabular}{cc}
		\toprule
		\textbf{Goal Position} & \textbf{Coordinates} \\
		\midrule
		Goal Position 1 & [1.2, 1.2] \\
		Goal Position 2 & [1.2, -1.2] \\
		Goal Position 3 & [-1.2, 1.2] \\
		Goal Position 4 & [-1.2, -1.2] \\
		Goal Position 5 & [1.2, 0] \\
		Goal Position 6 & [0, 1.2] \\
		Goal Position 7 & [-1.2, 0] \\
		Goal Position 8 & [0, -1.2] \\
		\bottomrule
	\end{tabular}
	
	\label{tab:goal_positions}
\end{table}

The start point is a fixed point as: [-0.2, -0.5]. To enhance the stability of 
the agent's initial position and reduce the risk of misalignment in real-world 
deployment, we introduce uniform noise to the starting coordinates. This 
approach aims to minimize data uncertainty caused by position discrepancies. By 
adding noise uniformly in the range $[-0.05, 0.05]$, we can achieve the desired 
effect.
Let:
\begin{itemize}
	\item \(\mathbf{a}_{\text{location}}\) be the agent's position
	\item \( U(a, b) \) denote a uniform distribution in the range $[a, b]$
\end{itemize}
The updated position formula is:
\begin{equation}
\mathbf{a}_{\text{location}} = \begin{bmatrix}
-0.2 + U(-0.05, 0.05) \\
-0.5 + U(-0.05, 0.05)
\end{bmatrix}
\end{equation}
This method ensures that the uniform noise maintains the desired variability 
without introducing bias \cite{levine2016}.

All in all, the state dimension is 49, and their shape and space are shown in 
\autoref{table:state}.

\begin{table}[htbp]
	\centering
	\caption{Noteable, Limo cannot make the control duration correctly if the duration is not in this time range.}
	\resizebox{\columnwidth}{!}{
		\begin{tabular}{llcl}
			\toprule
			\multicolumn{4}{c}{State details} \\
			\midrule
			Name                  & Shape                  &      Space           & Annotation                 \\
			\midrule
			Limo Position        & $[2, ]$                   & $[-1.5, 1.5]$      & in (X, Y), (meters) \\
			Goal Position        & $[2, ]$                   & $[-1.5, 1.5]$      &  in (X, Y), (meters)   \\
			Linear Velocity      & $[1, ]$                   & $[-1.0, 1.0]$      &  \\
			Steering Angle       & $[1, ]$                   & $[-1.0, 1.0]$      &  \\
			Control duration     & $[1, ]$                   & $[0.02, 0.5]$      & in Seconds                    \\
			Previous linear velocity & $[1, ]$               & $[-1.0, 1.0]$      &                     \\
			Angular velocity     & $[1, ]$                   & $[-1.0, 1.0]$      &                    \\
			20 radar point positions  & $[40, ]$             & $[-1.5, 1.5]$      & reshape from $[20, 2]$, in (X, Y) \\
			\bottomrule
		\end{tabular}
	}
	\label{table:state}
	
\end{table}


\begin{table}[htbp]
	\centering
	\caption{Noteable, Limo cannot make the control duration correctly if the duration is not in this time range.}
	\resizebox{\columnwidth}{!}{
		\begin{tabular}{llcl}
			\toprule
			\multicolumn{4}{c}{Action details} \\
			\midrule
			Name                  & Shape                  &      Space           & Annotation                 \\
			\midrule
			Control duration     & $[1, ]$                   & $[0.02, 0.5]$      & in Seconds                    \\
			Linear velocity      & $[1, ]$               & $[-1.0, 1.0]$      &                     \\
			Angular velocity     & $[1, ]$                   & $[-1.0, 1.0]$      &                    \\
			\bottomrule
		\end{tabular}
	}
	\label{table:action}
	
\end{table}


\begin{table}[htbp]
	\centering
	\caption{Reward Value Settings for Limo Environment}
	\resizebox{\columnwidth}{!}{
		\begin{tabular}{lcl}
			\toprule
			\multicolumn{3}{c}{Reward Settings} \\
			\midrule
			Name                       & Value              & Annotation                 \\
			\midrule
			$\text{cross\_punish}$     & $-30.0$            & cross with the enclosed regions \\
			$\text{success\_reward}$   & $500.0$            & success to reach the goal       \\
			$\text{dead\_punish}$      & $-100.0$           & go out of the map               \\
			$\delta$                   & $0.2$              & if limo close enough to a point, in meters                      \\
			\bottomrule
	\end{tabular}}
	\label{table:reward}
\end{table}

\section{System details}
\label{appendix:os}


\begin{table}[htbp]
	\centering
	\caption{Training PC Details}
	\begin{tabular}{ll}
		\toprule
		\multicolumn{2}{c}{PC Key Softwares' Ecosystem} \\
		\midrule
		Name                       & Value            \\
		\midrule
		Nvidia Driver Version      & 450.67           \\
		Cuda Version               & 11.8             \\
		cuDNN Version              & 8.9.7            \\
		Python Version             & 3.8              \\
		Torch Version              & 2.1.0            \\
		ONNX Version               & 1.13.1           \\
		Gymnasium Version          & 0.29.1           \\
		\bottomrule
	\end{tabular}
	\label{table:pc}
\end{table}

\begin{table}[htbp]
	\centering
	\caption{Agilex Limo Details}
	\begin{tabular}{ll}
		\toprule
		\multicolumn{2}{c}{Agilex Limo Key Softwares' Ecosystem} \\
		\midrule
		Name                       & Value            \\
		\midrule
		Nvidia JetPack Version     & 4.6.4            \\
		Cuda Version               & 10.2\_r440      \\
		cuDNN Version              & 8.2.1            \\
		TensorRT Version           & 8.2.1.3          \\
		Python Version             & 3.6              \\
		Pycuda Version             & 2020.1           \\
		ONNX Version               & 1.13.1           \\
		Limo Controller Version    & 2.4              \\
		\bottomrule
	\end{tabular}
	\label{table:limo}
\end{table}

Our experiments used the Agilex Limo equipped with a Jetson Nano running 
JetPack version 4.6.4. While newer versions such as 6.0+ are available, 
compatibility and support limitations necessitated our use of JetPack 4.6.4. 
Specifically, the older versions of cuDNN and TensorRT required for our setup 
are no longer available for direct download, although this may affect 
reproducibility. To assist others in replicating our results, we have provided 
custom-built support packages on our 
\href{https://github.com/alpaficia/MOSEAC_Limo}{GitHub}.

\section{Statistics for Energy and Time Consumptions}
\label{appendix:statistic_result}

Statistic results for the MOSEAC and SEAC on energy and time consumptions are 
shown in \autoref{tab:descriptives_tro}.

\begin{table}[h]
	\centering
	\caption{Descriptives}
	\label{tab:descriptives_tro}
	{
		\begin{tabular}{lrrrrr}
			\toprule
			& N & Mean & SD & SE & COV \\
			\cmidrule[0.4pt]{1-6}
			MOSEAC\_Energy & 100 & 10.130 & 4.733 & 0.473 & 0.467 \\
			SEAC\_Energy & 100 & 11.660 & 4.740 & 0.474 & 0.407 \\
			MOSEAC\_Time & 100 & 4.341 & 2.038 & 0.204 & 0.469 \\
			SEAC\_Time & 100 & 4.456 & 2.044 & 0.204 & 0.459 \\
			\bottomrule
		\end{tabular}
	}
\end{table}


\begin{table}[h]
	\centering
	\caption{Test of Normality (Shapiro-Wilk)}
	\label{tab:testOfNormality(Shapiro-Wilk)_tro}
	\begin{tabular}{lrrrr}
		\toprule
		&  &  & W & p \\
		\midrule
		MOSEAC\_energy & - & SEAC\_energy & 0.916 & $< 0.001$ \\
		MOSEAC\_time & - & SEAC\_time & 0.993 & 0.894 \\
		\bottomrule
	\end{tabular}
\end{table}


\begin{table}[htbp]
	\centering
	\caption{Paired Samples T-Test}
	\label{tab:pairedSamplesT-Test_tro}
	\resizebox{\linewidth}{!}{
		\begin{tabular}{lrrrrr}
			\toprule
			Measure 1 &  & Measure 2 & W & z & p \\
			\cmidrule[0.4pt]{1-6}
			MOSEAC\_Energy & - & SEAC\_Energy & 35.000 & -7.904 & \textless 0.001 \\
			MOSEAC\_Time & - & SEAC\_Time & 502.000 & -6.956 & \textless 0.001 \\
			\bottomrule
		\end{tabular}
	}
\end{table}

\section{Statistics for Compute Resources}
\label{appendix:compute_statistic}


\begin{table}[h]
	\centering
	\caption{Descriptives}
	\label{tab:descriptives_comp}
	\resizebox{\linewidth}{!}{
		\begin{tabular}{lrrrrr}
			\toprule
			& N & Mean & SD & SE & COV \\
			\cmidrule[0.4pt]{1-6}
			MOSEAC CPU usage (\%) & 88 & 11.400 & 0.370 & 0.131 & 0.032 \\
			SAC 10 Hz CPU usage (\%) & 88 & 16.800 & 0.400 & 0.141 & 0.024 \\
			SAC 60 Hz CPU usage (\%) & 88 & 31.413 & 1.298 & 0.459 & 0.041 \\
			MOSEAC GPU usage (\%) & 88 & 2.800 & 0.283 & 0.100 & 0.101 \\
			SAC 10 Hz GPU usage (\%) & 88 & 13.787 & 0.242 & 0.085 & 0.018 \\
			SAC 60 Hz GPU usage (\%) & 88 & 27.863 & 0.463 & 0.164 & 0.017 \\
			\bottomrule
	\end{tabular}}
\end{table}


\begin{table}[h]
	\centering
	\caption{Test of Normality (Shapiro-Wilk)}
	\label{tab:testOfNormality(Shapiro-Wilk)_comp}
	\resizebox{\linewidth}{!}{
	\begin{tabular}{lrrr}
		\toprule
		& W & p \\
		\cmidrule[0.4pt]{1-4}
		Measure 1 & Measure 2 & W & p \\
		\cmidrule[0.4pt]{1-4}
		MOSEAC CPU usage (\%) & SAC 10 Hz CPU usage (\%) & 0.947 & 0.685 \\
		& SAC 60 Hz CPU usage (\%) & 0.960 & 0.806 \\
		MOSEAC GPU usage (\%) & SAC 10 Hz GPU usage (\%) & 0.835 & 0.067 \\
		& SAC 60 Hz GPU usage (\%) & 0.901 & 0.296 \\
		\bottomrule
	\end{tabular}}
\end{table}


\begin{table}[h]
	\centering
	\caption{Paired Samples T-Test}
	\label{tab:pairedSamplesT-Test_comp}
	\resizebox{\linewidth}{!}{
		\begin{tabular}{lrrrrr}
			\toprule
			Measure 1 &  & Measure 2 & W & z  & p \\
			\cmidrule[0.4pt]{1-6}
			MOSEAC CPU usage (\%) & - & SAC 10 Hz CPU usage (\%) & - & -2.521 & 0.004 \\
			& - & SAC 60 Hz CPU usage (\%) & - & -2.521 & 0.004 \\
			MOSEAC GPU usage (\%) & - & SAC 10 Hz GPU usage (\%) & - & -2.521 & 0.007 \\
			& - & SAC 60 Hz GPU usage (\%) & - & -2.521 & 0.007 \\
			\bottomrule
	\end{tabular}}
\end{table}

\bibliographystyle{IEEEtran}
\bibliography{reference.bib}

\begin{thebibliography}{10}
\providecommand{\url}[1]{#1}
\csname url@samestyle\endcsname
\providecommand{\newblock}{\relax}
\providecommand{\bibinfo}[2]{#2}
\providecommand{\BIBentrySTDinterwordspacing}{\spaceskip=0pt\relax}
\providecommand{\BIBentryALTinterwordstretchfactor}{4}
\providecommand{\BIBentryALTinterwordspacing}{\spaceskip=\fontdimen2\font plus
\BIBentryALTinterwordstretchfactor\fontdimen3\font minus
  \fontdimen4\font\relax}
\providecommand{\BIBforeignlanguage}[2]{{%
\expandafter\ifx\csname l@#1\endcsname\relax
\typeout{** WARNING: IEEEtran.bst: No hyphenation pattern has been}%
\typeout{** loaded for the language `#1'. Using the pattern for}%
\typeout{** the default language instead.}%
\else
\language=\csname l@#1\endcsname
\fi
#2}}
\providecommand{\BIBdecl}{\relax}
\BIBdecl

\bibitem{liu2021deep}
R.~Liu, F.~Nageotte, P.~Zanne, M.~de~Mathelin, and B.~Dresp-Langley, ``Deep
  reinforcement learning for the control of robotic manipulation: a focussed
  mini-review,'' \emph{Robotics}, vol.~10, no.~1, p.~22, 2021.

\bibitem{ibarz2021train}
J.~Ibarz, J.~Tan, C.~Finn, M.~Kalakrishnan, P.~Pastor, and S.~Levine, ``How to
  train your robot with deep reinforcement learning: lessons we have learned,''
  \emph{The International Journal of Robotics Research}, vol.~40, no. 4-5, pp.
  698--721, 2021.

\bibitem{akalin2021reinforcement}
N.~Akalin and A.~Loutfi, ``Reinforcement learning approaches in social
  robotics,'' \emph{Sensors}, vol.~21, no.~4, p. 1292, 2021.

\bibitem{singh2022reinforcement}
B.~Singh, R.~Kumar, and V.~P. Singh, ``Reinforcement learning in robotic
  applications: a comprehensive survey,'' \emph{Artificial Intelligence
  Review}, vol.~55, no.~2, pp. 945--990, 2022.

\bibitem{majumdar2021paracosm}
R.~Majumdar, A.~Mathur, M.~Pirron, L.~Stegner, and D.~Zufferey, ``Paracosm: A
  test framework for autonomous driving simulations,'' in \emph{International
  Conference on Fundamental Approaches to Software Engineering}.\hskip 1em plus
  0.5em minus 0.4em\relax Springer International Publishing Cham, 2021, pp.
  172--195.

\bibitem{bregu2016reactive}
E.~Bregu, N.~Casamassima, D.~Cantoni, L.~Mottola, and K.~Whitehouse, ``Reactive
  control of autonomous drones,'' in \emph{Proceedings of the 14th Annual
  International Conference on Mobile Systems, Applications, and Services},
  2016, pp. 207--219.

\bibitem{karimi2023dynamic}
A.~Karimi, J.~Jin, J.~Luo, A.~R. Mahmood, M.~Jagersand, and S.~Tosatto,
  ``Dynamic decision frequency with continuous options,'' in \emph{2023
  IEEE/RSJ International Conference on Intelligent Robots and Systems
  (IROS)}.\hskip 1em plus 0.5em minus 0.4em\relax IEEE, 2023, pp. 7545--7552.

\bibitem{wang2024deployable}
D.~Wang and G.~Beltrame, ``Deployable reinforcement learning with variable
  control rate,'' \emph{arXiv preprint arXiv:2401.09286}, 2024.

\bibitem{wang2024reinforcement}
------, ``Reinforcement learning with elastic time steps,'' \emph{arXiv
  preprint arXiv:2402.14961}, 2024.

\bibitem{wang2024moseac}
------, ``Moseac: Streamlined variable time step reinforcement learning,''
  \emph{arXiv preprint arXiv:2406.01521}, 2024.

\bibitem{agilex_limo}
{AgileX Robotics}, ``Agilex limo - multi-modal mobile robot with ai modules,''
  \url{https://www.globenewswire.com/}, accessed: [date].

\bibitem{lajoie2020door}
P.-Y. Lajoie, B.~Ramtoula, Y.~Chang, L.~Carlone, and G.~Beltrame, ``Door-slam:
  Distributed, online, and outlier resilient slam for robotic teams,''
  \emph{IEEE Robotics and Automation Letters}, vol.~5, no.~2, pp. 1656--1663,
  2020.

\bibitem{wan2020cognitive}
S.~Wan, Z.~Gu, and Q.~Ni, ``Cognitive computing and wireless communications on
  the edge for healthcare service robots,'' \emph{Computer Communications},
  vol. 149, pp. 99--106, 2020.

\bibitem{bouteiller2021reinforcement}
Y.~Bouteiller, S.~Ramstedt, G.~Beltrame, C.~Pal, and J.~Binas, ``Reinforcement
  learning with random delays,'' in \emph{International conference on learning
  representations}, 2021.

\bibitem{amin2020locally}
S.~Amin, M.~Gomrokchi, H.~Aboutalebi, H.~Satija, and D.~Precup, ``Locally
  persistent exploration in continuous control tasks with sparse rewards,''
  \emph{arXiv preprint arXiv:2012.13658}, 2020.

\bibitem{park2021time}
S.~Park, J.~Kim, and G.~Kim, ``Time discretization-invariant safe action
  repetition for policy gradient methods,'' \emph{Advances in Neural
  Information Processing Systems}, vol.~34, pp. 267--279, 2021.

\bibitem{sharma2017learning}
S.~Sharma, A.~Srinivas, and B.~Ravindran, ``Learning to repeat: Fine grained
  action repetition for deep reinforcement learning,'' \emph{arXiv preprint
  arXiv:1702.06054}, 2017.

\bibitem{metelli2020control}
A.~M. Metelli, F.~Mazzolini, L.~Bisi, L.~Sabbioni, and M.~Restelli, ``Control
  frequency adaptation via action persistence in batch reinforcement
  learning,'' in \emph{International Conference on Machine Learning}.\hskip 1em
  plus 0.5em minus 0.4em\relax PMLR, 2020, pp. 6862--6873.

\bibitem{lee2020reinforcement}
J.~Lee, B.-J. Lee, and K.-E. Kim, ``Reinforcement learning for control with
  multiple frequencies,'' \emph{Advances in Neural Information Processing
  Systems}, vol.~33, pp. 3254--3264, 2020.

\bibitem{chen2021varlenmarl}
Y.~Chen, H.~Wu, Y.~Liang, and G.~Lai, ``Varlenmarl: A framework of
  variable-length time-step multi-agent reinforcement learning for cooperative
  charging in sensor networks,'' in \emph{2021 18th Annual IEEE International
  Conference on Sensing, Communication, and Networking (SECON)}.\hskip 1em plus
  0.5em minus 0.4em\relax IEEE, 2021, pp. 1--9.

\bibitem{even2003action}
E.~Even-Dar, S.~Mannor, and Y.~Mansour, ``Action elimination and stopping
  conditions for reinforcement learning,'' in \emph{Proceedings of the 20th
  International Conference on Machine Learning (ICML-03)}, 2003, pp. 162--169.

\bibitem{zhao2023reinforcement}
S.~Zhao, T.~Zheng, D.~Sui, J.~Zhao, and Y.~Zhu, ``Reinforcement learning based
  variable damping control of wearable robotic limbs for maintaining astronaut
  pose during extravehicular activity,'' \emph{Frontiers in Neurorobotics},
  vol.~17, p. 1093718, 2023.

\bibitem{Gottipati2020MaxReward}
\BIBentryALTinterwordspacing
N.~Gottipati \emph{et~al.}, ``To the max: Reinventing reward in reinforcement
  learning,'' \emph{arXiv preprint arXiv:2402.01361}, 2020. [Online].
  Available: \url{https://ar5iv.labs.arxiv.org/html/2402.01361}
\BIBentrySTDinterwordspacing

\bibitem{sutton2018reinforcement}
R.~S. Sutton and A.~G. Barto, \emph{Reinforcement learning: An
  introduction}.\hskip 1em plus 0.5em minus 0.4em\relax MIT press, 2018.

\bibitem{dietterich2000hierarchical}
T.~G. Dietterich, ``Hierarchical reinforcement learning with the maxq value
  function decomposition,'' \emph{Journal of artificial intelligence research},
  vol.~13, pp. 227--303, 2000.

\bibitem{li2019hierarchical}
S.~Li, R.~Wang, M.~Tang, and C.~Zhang, ``Hierarchical reinforcement learning
  with advantage-based auxiliary rewards,'' \emph{Advances in Neural
  Information Processing Systems}, vol.~32, 2019.

\bibitem{kacem2002pareto}
I.~Kacem, S.~Hammadi, and P.~Borne, ``Pareto-optimality approach for flexible
  job-shop scheduling problems: hybridization of evolutionary algorithms and
  fuzzy logic,'' \emph{Mathematics and computers in simulation}, vol.~60, no.
  3-5, pp. 245--276, 2002.

\bibitem{monfared2021pareto}
M.~S. Monfared, S.~E. Monabbati, and A.~R. Kafshgar, ``Pareto-optimal
  equilibrium points in non-cooperative multi-objective optimization
  problems,'' \emph{Expert Systems with Applications}, vol. 178, p. 114995,
  2021.

\bibitem{haarnoja2018soft}
T.~Haarnoja, A.~Zhou, K.~Hartikainen, G.~Tucker, S.~Ha, J.~Tan, V.~Kumar,
  H.~Zhu, A.~Gupta, P.~Abbeel \emph{et~al.}, ``Soft actor-critic algorithms and
  applications,'' \emph{arXiv preprint arXiv:1812.05905}, 2018.

\bibitem{banach1922operations}
S.~Banach, ``Sur les op{\'e}rations dans les ensembles abstraits et leur
  application aux {\'e}quations int{\'e}grales,'' \emph{Fundamenta
  mathematicae}, vol.~3, no.~1, pp. 133--181, 1922.

\bibitem{konda1999actor}
V.~Konda and J.~Tsitsiklis, ``Actor-critic algorithms,'' \emph{Advances in
  neural information processing systems}, vol.~12, 1999.

\bibitem{optitrack}
{OptiTrack}, ``Optitrack - motion capture systems,''
  \url{https://www.optitrack.com}, accessed: [date].

\bibitem{rybczak2024survey}
M.~Rybczak, N.~Popowniak, and A.~Lazarowska, ``A survey of machine learning
  approaches for mobile robot control,'' \emph{Robotics}, vol.~13, no.~1,
  p.~12, 2024.

\bibitem{imece2022}
Authors, ``Supervised and unsupervised deep learning applications for visual
  slam in robotics,'' in \emph{IMECE}.\hskip 1em plus 0.5em minus 0.4em\relax
  ASME, 2022, p. V003T04A010.

\bibitem{vaswani2023attention}
A.~Vaswani, N.~Shazeer, N.~Parmar, J.~Uszkoreit, L.~Jones, A.~N. Gomez,
  L.~Kaiser, and I.~Polosukhin, ``Attention is all you need,'' 2023.

\bibitem{impact_distribution_shift_2024}
\BIBentryALTinterwordspacing
Authors, ``Assessing the impact of distribution shift on reinforcement learning
  performance,'' \emph{arXiv preprint arXiv:2402.03590}, 2024. [Online].
  Available: \url{https://arxiv.org/abs/2402.03590}
\BIBentrySTDinterwordspacing

\bibitem{katakura2023rl_dynamic_state_space}
\BIBentryALTinterwordspacing
M.~Katakura \emph{et~al.}, ``Reinforcement learning model with dynamic state
  space tested on target search tasks for monkeys: Extension to learning task
  events,'' \emph{Frontiers in Robotics and AI}, 2023. [Online]. Available:
  \url{https://www.frontiersin.org/articles/10.3389/frobt.2023.00856/full}
\BIBentrySTDinterwordspacing

\bibitem{nvidia_jetson_nano}
``Jetson nano module,''
  \url{https://developer.nvidia.com/embedded/jetson-nano}, 2019.

\bibitem{paszke2019pytorch}
\BIBentryALTinterwordspacing
A.~Paszke, S.~Gross, F.~Massa, A.~Lerer, J.~Bradbury, G.~Chanan, T.~Killeen,
  Z.~Lin, N.~Gimelshein, L.~Antiga, A.~Desmaison, A.~Kopf, E.~Yang, Z.~DeVito,
  M.~Raison, A.~Tejani, S.~Chilamkurthy, B.~Steiner, L.~Fang, J.~Bai, and
  S.~Chintala, ``Pytorch: An imperative style, high-performance deep learning
  library,'' in \emph{Advances in Neural Information Processing Systems},
  vol.~32, 2019, pp. 8024--8035. [Online]. Available:
  \url{https://proceedings.neurips.cc/paper/2019/hash/bdbca288fee7f92f2bfa9f7012727740-Abstract.html}
\BIBentrySTDinterwordspacing

\bibitem{towers_gymnasium_2023}
\BIBentryALTinterwordspacing
M.~Towers, J.~K. Terry, A.~Kwiatkowski, J.~U. Balis, G.~d. Cola, T.~Deleu,
  M.~Goulão, A.~Kallinteris, A.~KG, M.~Krimmel, R.~Perez-Vicente, A.~Pierré,
  S.~Schulhoff, J.~J. Tai, A.~T.~J. Shen, and O.~G. Younis, ``Gymnasium,'' Mar.
  2023. [Online]. Available: \url{https://zenodo.org/record/8127025}
\BIBentrySTDinterwordspacing

\bibitem{tensorrt}
``Tensorrt,'' \url{https://developer.nvidia.com/tensorrt}, 2021.

\bibitem{quigley2009ros}
M.~Quigley, K.~Conley, B.~Gerkey, J.~Faust, T.~Foote, J.~Leibs, R.~Wheeler, and
  A.~Y. Ng, ``Ros: an open-source robot operating system,'' in \emph{ICRA
  workshop on open source software}, vol.~3, no. 3.2.\hskip 1em plus 0.5em
  minus 0.4em\relax Kobe, Japan, 2009, p.~5.

\bibitem{ros2}
S.~Macenski, T.~Foote, B.~Gerkey, C.~Lalancette, and W.~Woodall, ``Ros 2:
  Towards a performance-centric and real-time robotics framework,'' in
  \emph{Robotics: Science and Systems (RSS) Workshop on Real-time and
  Performance in Robotic Systems}, Online, 2020.

\bibitem{docker}
D.~Inc., \emph{Docker: Open Platform for Developing, Shipping, and Running
  Applications}, 2013, available at \url{https://docs.docker.com/}.

\bibitem{ryan2004review}
H.~Ryan, M.~Paglione, and S.~Green, ``Review of trajectory accuracy methodology
  and comparison of error measurement metrics,'' in \emph{AIAA Guidance,
  Navigation, and Control Conference and Exhibit}, 2004, p. 4787.

\bibitem{muller2007dynamic}
M.~M{\"u}ller, ``Dynamic time warping,'' \emph{Information retrieval for music
  and motion}, pp. 69--84, 2007.

\bibitem{pareto1971manual}
V.~Pareto, \emph{Manual of Political Economy}.\hskip 1em plus 0.5em minus
  0.4em\relax New York: Augustus M. Kelley, 1971.

\bibitem{deb2002fast}
K.~Deb, A.~Pratap, S.~Agarwal, and T.~Meyarivan, ``A fast and elitist
  multiobjective genetic algorithm: Nsga-ii,'' \emph{IEEE transactions on
  evolutionary computation}, vol.~6, no.~2, pp. 182--197, 2002.

\bibitem{riedmiller2018}
\BIBentryALTinterwordspacing
M.~Riedmiller, R.~Hafner, T.~Lampe, M.~Neunert, J.~Degrave, T.~van~de Wiele,
  V.~Mnih, and N.~Heess, ``Learning by playing - solving sparse reward tasks
  from scratch,'' in \emph{Proceedings of the 35th International Conference on
  Machine Learning}, 2018. [Online]. Available:
  \url{https://arxiv.org/abs/1802.09464}
\BIBentrySTDinterwordspacing

\bibitem{bellemare2016}
\BIBentryALTinterwordspacing
M.~G. Bellemare, S.~Srinivasan, G.~Ostrovski, T.~Schaul, D.~Saxton, and
  R.~Munos, ``Unifying count-based exploration and intrinsic motivation,''
  \emph{arXiv preprint arXiv:1704.08302}, 2017. [Online]. Available:
  \url{https://arxiv.org/abs/1704.08302}
\BIBentrySTDinterwordspacing

\bibitem{levine2016}
\BIBentryALTinterwordspacing
S.~Levine, C.~Finn, T.~Darrell, and P.~Abbeel, ``End-to-end training of deep
  visuomotor policies,'' \emph{The Journal of Machine Learning Research},
  vol.~17, no.~1, pp. 1334--1373, 2016. [Online]. Available:
  \url{https://arxiv.org/abs/1604.07316}
\BIBentrySTDinterwordspacing

\end{thebibliography}

\begin{IEEEbiography}[{\includegraphics[width=1in,height=1.25in,clip,
  		keepaspectratio]{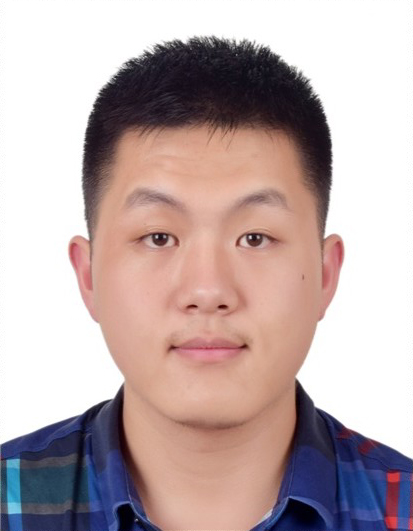}}]{Dong Wang}
  (Member, IEEE) received his bachelor's degree in electronic engineering from the 
  School of Aviation, Northwestern Polytechnical University (NWPU), Xi'an, China, 
  in 2017. He is pursuing his Ph.D. in the Department of Software Engineering at 
  Polytechnique Montreal, Montreal, Canada. His research interests include 
  reinforcement learning, computer vision, and robotics.
\end{IEEEbiography}

\vspace{11pt}
\begin{IEEEbiography}[{\includegraphics[width=1in,height=1.25in,clip,
  		keepaspectratio]{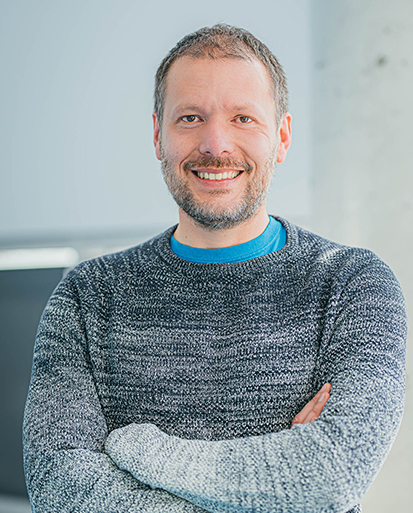}}]{Giovanni Beltrame}
  Giovanni Beltrame (Senior Member, IEEE) received
  the Ph.D. degree in computer engineering from Po-
  litecnico di Milano, Milan, Italy, in 2006.
  He worked as a Microelectronics Engineer with the
  European Space Agency, Paris, France, on a number
  of projects, spanning from radiation tolerant systems
  to computer-aided design. Since 2010, he has been the
  Professor with the Computer and Software Engineer-
  ing Department, Polytechnique Montreal, Montreal,
  QC, Canada, where he directs the MIST Lab. He
  has authored or coauthored more than 150 papers in
  international journals and conferences. His research interests include modeling
  and design of embedded systems, artificial intelligence, and robotics.
\end{IEEEbiography}

\vfill

\end{document}